%% file: main.tex
\documentclass[10pt,journal,compsoc]{IEEEtran}

\ifCLASSOPTIONcompsoc
  \usepackage[nocompress]{cite}
\else
  \usepackage{cite}
\fi

\ifCLASSINFOpdf
    \usepackage[pdftex]{graphicx}
    \usepackage{float}
\else
\fi

\usepackage{caption2}
\DeclareCaptionLabelFormat{cont}{#1~#2\alph{ContinuedFloat}}
\usepackage{color}
\usepackage{subfigure}
\usepackage{amsmath}
\usepackage{amsfonts}
\usepackage{algpseudocode}
\usepackage{multirow}
\usepackage{multicol}
\usepackage{array}
\usepackage{xcolor}
\usepackage{caption2}
\usepackage{subfigure}
\usepackage{stfloats}
\usepackage{booktabs}
\usepackage{placeins}
\usepackage[colorlinks,linkcolor=blue]{hyperref}

\newcommand{\rao}[1]{{\color{black}#1}}

\newcommand{\gl}[1]{{\color{black}#1}}
\newcommand{\glc}[1]{{\color{black}#1}}

\newcommand{\YL}[1]{{\color{black}#1}}

\begin{document}
\title{RISA-Net: Rotation-Invariant Structure-Aware Network for Fine-Grained 3D Shape Retrieval}

\author{Rao~Fu,%
        ~Jie~Yang,%
        ~Jiawei~Sun,%
        ~Fang-Lue~Zhang,
        ~Yu-Kun~Lai
        and
        ~Lin~Gao
\IEEEcompsocitemizethanks{
\IEEEcompsocthanksitem R. Fu, J. Yang and L. Gao are with the Beijing Key Laboratory of Mobile Computing and Pervasive Device, Institute of Computing Technology, Chinese Academy of Sciences, Beijing, China and also with University of Chinese Academy of Sciences, Beijing, China.\protect\\
E-mail: furao17@mails.ucas.ac.cn, \{jieyang, lingao\}@ict.ac.cn
\IEEEcompsocthanksitem  J. Sun is with University of Chinese Academy of Science, Beijing, China.\protect
\IEEEcompsocthanksitem F. Zhang is with School of Engineering and Computer Science, Victoria University of Wellington, New Zealand.\protect\\
E-mail:fanglue.zhang@vuw.ac.nz
\IEEEcompsocthanksitem Y.-K Lai is with School of Computer Science and Informatics, Cardiff University, Wales, UK.\protect ~E-mail:LaiY4@cardiff.ac.uk
}%
\thanks{Manuscript received September 25, 2020.}}

\markboth{Journal of \LaTeX\ Class Files,~Vol.~XX, No.~X, August~20XX}%
{Fu \MakeLowercase{\textit{et al.}}: RISA-Net: Rotation-Invariant Structure-Aware Network for Fine-Grained 3D Shape Retrieval}

\IEEEtitleabstractindextext{%
\begin{abstract}
Fine-grained 3D shape retrieval aims to retrieve 3D shapes similar to a query shape in a repository with models belonging to the same class, which requires shape descriptors to be capable of representing detailed geometric information to discriminate shapes with globally similar structures. Moreover, 3D objects can be placed with arbitrary position and orientation in real-world applications, which further requires shape descriptors to be robust to rigid transformations. The shape descriptions  used in existing 3D shape retrieval systems fail to meet the above two criteria. In this paper, we introduce a novel deep architecture, RISA-Net, which learns rotation invariant 3D shape descriptors that are capable of encoding fine-grained geometric information and structural information, and thus achieve accurate results on the task of fine-grained 3D object retrieval. RISA-Net extracts a set of compact and detailed geometric features part-wisely and discriminatively estimates the contribution of each semantic part to shape representation. 
Furthermore, our method is able to learn the importance of geometric and structural information of all the parts when generating the final compact latent feature of a 3D shape for fine-grained retrieval. 
We also build and publish a new 3D shape dataset with sub-class labels for validating the performance of fine-grained 3D shape retrieval methods. Qualitative and quantitative experiments show that our RISA-Net outperforms state-of-the-art methods on the fine-grained object retrieval task, demonstrating its  capability in geometric detail extraction. 
The code and dataset are available at: \href{https://github.com/IGLICT/RisaNET}{https://github.com/IGLICT/RisaNET}.
\end{abstract}

\begin{IEEEkeywords}
3D Shape Retrieval, Fine-grained Retrieval, Rotation-invariant Representation, Convolutional Neural Networks
\end{IEEEkeywords}}

\maketitle

\IEEEdisplaynontitleabstractindextext

\IEEEpeerreviewmaketitle

\input{tex/3DShapeRetrieval}

\ifCLASSOPTIONcaptionsoff
  \newpage
\fi

\bibliographystyle{IEEEtran}
\bibliography{IEEEabrv,mylib}

\begin{IEEEbiography}
[{\includegraphics[width=0.9in,height=1.25in,clip,keepaspectratio]{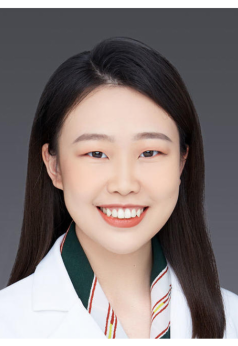}}]
{Rao Fu} is now a bachelor candidate in the University of Chinese Academy of Sciences, majoring in computer science. She has been a visiting researcher in University of California, San Diego. Her research interests include computer graphics, geometric processing and 3D computer vision.
\end{IEEEbiography}

\vspace{-2.0cm}
\begin{IEEEbiography}[{\includegraphics[width=1in,height=1.25in,clip,keepaspectratio]{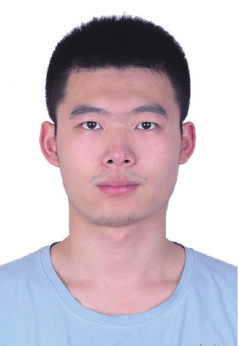}}]{Jie Yang}
received a bachelor's degree in mathematics from Sichuan University in 2016. He is currently a PhD candidate in the Institute of Computing Technology, Chinese Academy of Sciences. His research interests include computer graphics and geometric processing.
\end{IEEEbiography}

\vspace{-2.0cm}
\begin{IEEEbiography}
[{\includegraphics[width=1in,height=1.25in,clip,keepaspectratio]{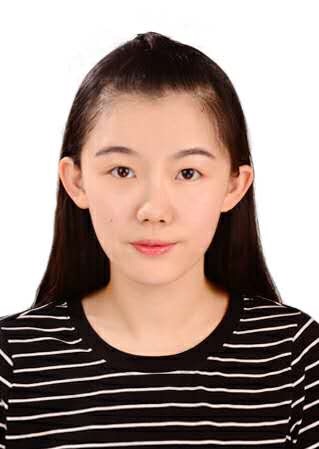}}]
{Jiawei Sun}
is now majoring in computer science. She is currently an undergraduate in  Beijing Jiaotong University. Her research interests include computer graphics and geometric processing.
\end{IEEEbiography}

\vspace{-2.0cm}
\begin{IEEEbiography}[{\includegraphics[width=1in,height=1.25in,clip,keepaspectratio]{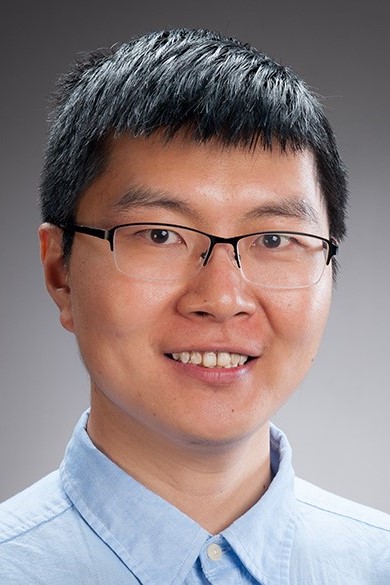}}]
{Fang-Lue Zhang}
is currently a lecturer with Victoria University of
Wellington, New Zealand. He received the Bachelors degree from Zhejiang University, Hangzhou, China, in 2009, and the Doctoral degree from Tsinghua University, Beijing, China, in 2015. His research interests include image and video editing, computer vision, and computer graphics. He is a member of IEEE and ACM. He received Victoria Early Career Research Excellence Award in 2019.
\end{IEEEbiography}

\vspace{-2.0cm}
\begin{IEEEbiography}[{\includegraphics[width=1in,height=1.25in,clip,keepaspectratio]{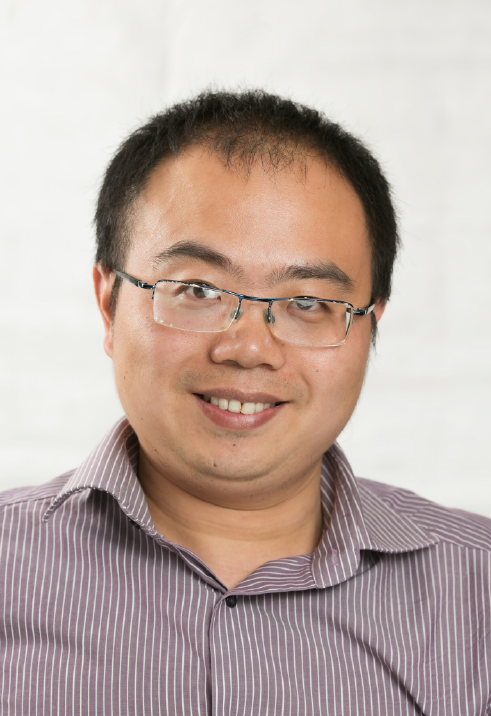}}]{Yu-Kun Lai}
received his bachelor's degree and PhD degree in computer science from
Tsinghua University in 2003 and 2008, respectively. He is currently a Professor in the School of Computer Science \& Informatics, Cardiff University. His research
interests include computer graphics, geometry processing, image processing and computer vision. He is on the editorial boards of \emph{Computer Graphics Forum} and \emph{The Visual Computer}.
\end{IEEEbiography}

\vspace{-2.0cm}
\begin{IEEEbiography}[{\includegraphics[width=1in,height=1.25in,clip,keepaspectratio]{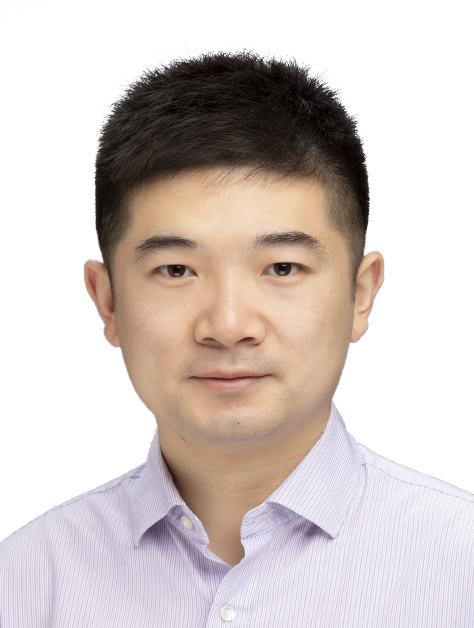}}]{Lin Gao}
received the bachelor's degree in mathematics from Sichuan University and the PhD degree in computer science from Tsinghua University. He is currently an Associate Professor at the Institute of Computing Technology, Chinese Academy of Sciences. He has been awarded the Newton Advanced Fellowship from the Royal Society and the AG young researcher award. His research interests include computer graphics and geometric processing. 
\end{IEEEbiography}

\newpage
\setcounter{section}{0}
\onecolumn
\input{tex/supp}

\end{document}

%% file: tex/3DShapeRetrieval.tex
\ifCLASSOPTIONcompsoc
\IEEEraisesectionheading{\section{Introduction}\label{sec:introduction}}
\IEEEPARstart{T}{he} recent advances in modeling, digitizing and visualizing 3D physical and virtual objects have led to an explosion in the number of available 3D models on the internet. Therefore, the need for effectively retrieving models from a shape repository has become an important and integral part in the research field of 3D shape analysis. The mainstream shape retrieval methods are content-based approaches, which use shape  descriptors to search for similar models. The extraction of shape descriptors with the capability of representing detailed geometric information and global structure is the fundamental task of such approaches. Also, as objects may be placed with arbitrary position and orientation in real-world applications, the requirement on the robustness of the descriptors to rigid transformations
increases the difficulty of the problem.
\begin{figure*}[t]
\begin{center}
\includegraphics[width=7.2in]{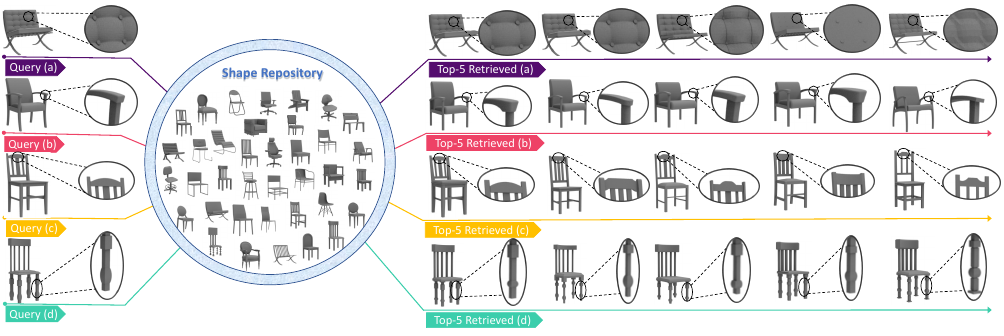}
\end{center}
   \caption{Four examples of the top-5 retrieval results given query models on the shape dataset with perturbed rotations within the chair category. Our method is able to capture geometric details, learn the importance of each part, and balance the contribution of structure and geometric information in fine-grained retrieval.}
   \label{fig:sample_general}
\end{figure*}

Although existing methods have been proposed to retrieve similar 3D shapes using their shape descriptors, these methods only perform well on coarse-level retrieval, where the shape search engines are asked to retrieve shapes with the same class label among different object classes. These classes normally have radically different %
overall shapes. %
Fine-grained object retrieval is more challenging compared with inter-class retrieval, where the search engine could be asked to %
find \rao{lounge chairs among various sub-classes of chairs, like Barcelona chairs, drafting chairs, slat chairs, etc.} Such a fine-grained retrieval task thus requires shape descriptors to be capable of encoding fine-grained properties very well, such as geometric details, because the shapes of different sub-classes in the repository could look similar.
Meanwhile, the robustness to rigid transformations, in particular rotations, should be better addressed in the research field of shape retrieval. 3D models in practical applications may not be aligned, but many retrieval methods are not fully invariant to cope with possible rigid transformations. 
\YL{Shape} aligning procedures are often required to apply these shape descriptors to \gl{3D shapes with \YL{arbitrary} rotations}\YL{, which is not only time consuming, but also less robust.}

To tackle the above challenges, we propose a \glc{Rotation-Invariant Structure-Aware Network \YL{(RISA-Net)} for fine-grained 3D shape retrieval} which extracts \YL{a} 3D shape descriptor that is robust to rigid \YL{transformation} and is \YL{also informative} to indicate fine-grained shape similarity. Our method is based on the following three observations. Firstly, in \YL{both} physical object manufacturing \YL{and} digital 3D shape modeling, assembling interchangeable components or parts has become a practical reality to avoid considerable, and often inconvenient user interactions in 3D object design ~\cite{Assembly_reference,Manufacturing_reference}. It is thus natural to see part-wise differences in 3D models constructed for different \YL{functionalities}. The parts of an object \YL{often vary in} importance, and \YL{therefore} contribute \YL{unequally} to discriminating different object categories. 
Therefore, we extract geometric information part-wisely and then use an attention mechanism to \YL{learn to} weight the \YL{contributions} of different parts to shape retrieval. Secondly, we need to ensure that the extracted descriptor can well encode the subtle difference among the same semantic part of different objects and is invariant to rigid transformations. Intuitively, a latent feature that can be precisely reconstructed to the whole shape must have comprised all detailed geometric information of the \YL{original} shape. Thus, we propose a variational autoencoder to encode reconstructive rotation-invariant geometric information: edge \YL{lengths} and dihedral \YL{angles}. Thirdly, the global structural information is also important in fine-grained object retrieval. Shapes of different sub-classes may share similar local geometry, but their structures could be different. Therefore, we propose a new paradigm to balance the importance of fine-grained geometric information and structural information. In particular, we incorporate the weighted rigid \YL{transformation} invariant structural information in shape descriptors through fully-connected layers. With the above components, our network is able to learn 3D shape descriptors that achieve a high accuracy when querying shapes with arbitrary poses against the objects of the same class.

In \YL{summary}, our main contributions are as follows:
\begin{enumerate}
    \item We propose a novel part-based deep model, RISA-Net, which learns fine-grained geometric information part-wisely that is robust to rigid \YL{transformation} and discriminatingly estimates the contribution of each part of the object to shape retrieval. It outperforms the state-of-the-art methods on fine-grained shape retrieval tasks.
    
    \item We propose a paradigm that balances structural and geometric information in shape discrimination by learning the weights of structural information and geometric information for the final shape latent feature generation.
    
    \item We build and publish a large 3D Shape dataset, RISA-Dataset, to evaluate fine-grained object retrieval methods quantitatively, which provides sub-class labels of all the $8,906$ 3D shapes.

\end{enumerate}

The following sections are arranged as follows: In Section~\ref{sec:related_work}, we briefly summarize the previous works. %
In Section~\ref{sec:method} we elaborate RISA-Net's structure through its two-stage feature representation learning and attention mechanism. Section~\ref{sec:Implementation} includes the fine-grained 3D shape retrieval dataset we constructed, implementation details and experiments to demonstrate the capability of RISA-Net. Finally we conclude our paper and bring up several limitations of our method that could be addressed in the future in Section~\ref{sec:conclusion}.

\else
\section{Introduction}
\label{sec:introduction}

\fi

\section{Related Work}
\label{sec:related_work}
3D shape retrieval is one of the fundamental tasks that \YL{discerns} the descriptive power of shape representation. In this section, we first review \YL{existing research} on 3D shape retrieval. Then we look into two-lines of state-of-the-art approaches for 3D shape representation: mesh-based representations and rotation-invariant representations\YL{, which are particularly relevant to our work. }

\subsection{3D Shape Retrieval}
Shape retrieval \YL{is fundamental to facilitate many} applications based on large-scale 3D shape repositories, such as shape modeling~\cite{xie2013sketch}, template-based deformation~\cite{uy2020deformation}, and scene modelling~\cite{xu2013sketch2scene}. These works retrieve a globally or locally similar shape of the target shape and use it as a component or template for shape \YL{modeling}, leveraging information provided by the shape repository. The essential component of shape retrieval is effective description of 3D shapes. Among hand-crafted shape descriptors, lightfield descriptors~\cite{chen2003visual} and spherical harmonic descriptors~\cite{kazhdan2003rotation} have been used to extract global 3D features, while shape distribution~\cite{osada2002shape}, heat kernel diffusion~\cite{sun2009concise}, predefined primitives~\cite{pratikakis2010learning} and bag-of-features~\cite{bronstein2011shape,redondo2012surfing,ohbuchi2010distance} are able to describe local information for partial shape retrieval, and aggregate local information for global shape retrieval. 
Recently, facilitated by progresses in deep neural networks, machine learning based methods are adopted to improve the descriptive power of 3D shape representations. Multi-view image-based approaches aim to aggregate features from multi-view images for shape representation, which \YL{aggregates} image features through pooling operations~\cite{su2015multi}, image matching~\cite{bai2016gift,kanezaki2018rotationnet}, or attention-based sequential view aggregation~\cite{han2018seqviews2seqlabels,han20193d2seqviews}. Instead of projecting shapes to multi-view images, Shi et al. \cite{shi2015deeppano} and Steve et al. \cite{esteves2018learning} proposed to project \YL{a} shape to a cylinder and \YL{a} unit sphere respectively, and learn features from their projection without extra aggregation operations. \YL{Other} methods focused on extracting features from point \YL{based} representations. \cite{furuya2016deep} extracted and aggregated local features from rotation-normalized point sets. \cite{zhou2018voxelnet} transformed the point set into a volumetric representation and introduced a voxel feature encoding layer for feature extraction.

While these methods perform well on large-scale 3D shape \YL{retrieval} benchmarks~\cite{li2012shrec,savva2016shrec16}, their representative power is limited to distinguishing shapes %
\YL{of different sub-classes within the same overall class,}
overlooking fine-grained shape features. Additionally, they focused on only geometric information or spatial information, without discerning the global semantic structure of objects. Observing their drawbacks in %
\YL{sub-class level}
retrieval, we propose RISA-Net that focuses on fine-grained shape retrieval, which is able to encode both geometric and structural features and \YL{weight} their importance. We also provide a dataset for \YL{quantitatively evaluating 
fine-grained shape retrieval.}

\subsection{Mesh-based Representations}
There \YL{have} been numerous studies \YL{that} investigate how to apply convolution \YL{operations} on 3D mesh-based models. Masci et al.~\cite{masci2015geodesic} \YL{were} the first to extract patches based on local polar coordinates and generalize convolution networks to non-Euclidean manifolds. Sinha et al.~\cite{sinha2016deep} used \YL{Convolutional Neural Networks (CNNs)} to transform a general mesh model into a ``geometry image'' that encodes local properties of shape surfaces. Anisotropic CNN (ACNN)~\cite{boscaini2016learning} adopted anisotropic diffusion kernels to construct patches to learn intrinsic correspondences. Monti et al.~\cite{monti2017geometric} further improved these ideas by parametrically constructing patch operators through vertex frequency analysis. \YL{Alternatively, methods were} reported in the literature to perform convolutional operations in the spectral domain. Boscaini et al.~\cite{boscaini2015learning} used windowed Fourier transform \YL{and} proposed localized spectral convolutional networks to conduct supervised local feature learning. Xie et al.~\cite{xie2016learned} learned \YL{a} binary spectral shape descriptor for 3D shape correspondence. 
\YL{Han et al.~\cite{han2016unsupervised} further} proposed a circle convolutional restricted Boltzmann machine (CCRBM) to learn 3D local features in an unsupervised manner. %
\YL{In follow-up work,}
a mesh convolutional restricted \YL{Boltzmann} machine was proposed~\cite{han2016mesh}, which hierarchically encodes the geometry spatiality of local regions through local function energy distribution of 3D meshes. Hanocka et al.\cite{hanocka2019meshcnn} brought up a network with unique convolution and pooling operations on the edges \YL{which connect adjacent mesh vertices}. Schult et al.~\cite{schult2020dualconvmesh} proposed a network that \YL{applies} geodesic and \YL{Euclidean} convolutional operations in parallel. These methods provide fundamental building blocks for deep learning methods for geometry processing.

A series of recent studies \YL{has} also indicated the effectiveness of generative models in the literature. The method introduced by Xie et al.~\cite{xie2015deepshape} is able to encode heat diffusion based 3D mesh descriptors in a multiscale manner with a set of deep discriminative auto-encoders. More recently, Litany et al.~\cite{litany2018deformable} proposed intrinsic \YL{Variational Autoencoders (VAEs)} for meshes, and applied them to shape completion. A \YL{deep} model that can learn both geometric and \YL{spatial} information was successfully established by Han et al.~\cite{han2018deep}. It learns various types of virtual word features \YL{in bag-of-words models} 
and encodes both global and local spatial information simultaneously in an unsupervised way. A recent study by Gao et al.~\cite{gao2019sdm} \YL{presented a deep generative model for high-quality shape generation through deformable mesh parts and hierarchical VAEs encoding both parts and their structure. }
Wu et al.~\cite{wu2019pq} proposed using sequence-to-sequence autoencoders to encode shape part-wisely.
Inspired by the above works, we propose to use generative models to automatically learn shape descriptors in the latent space.

\begin{figure*}[!htb]
\centering
\includegraphics[scale = 0.55]{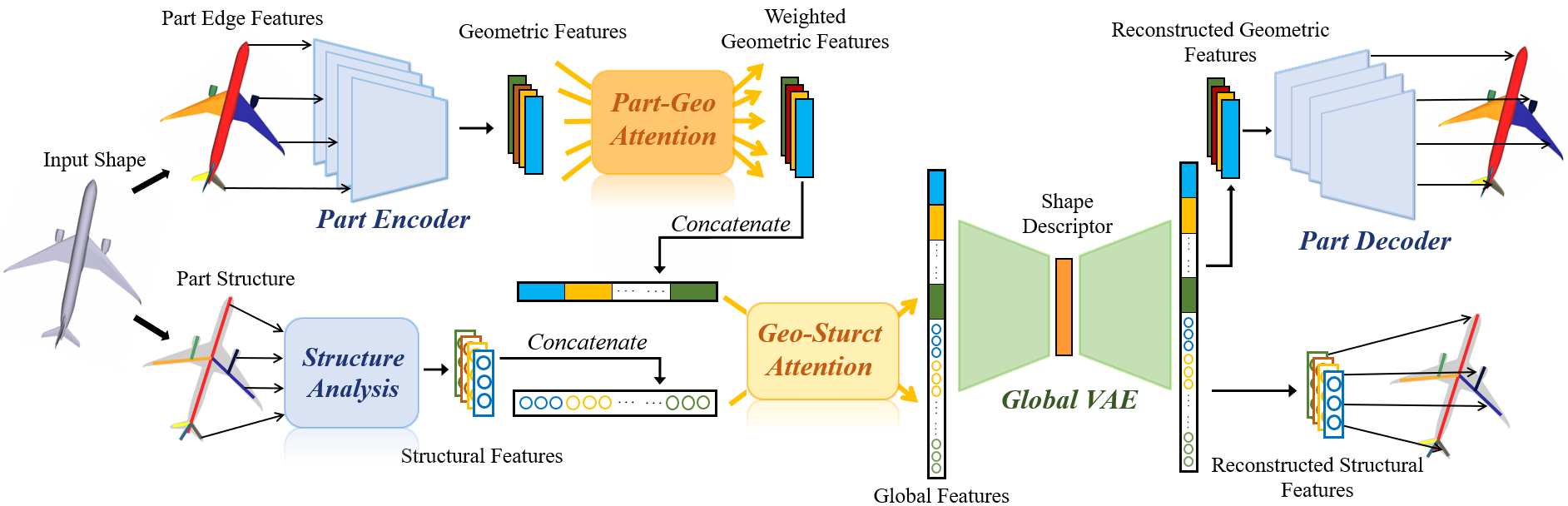}
   \caption{Pipeline of RISA-Net. From left to the right, i) our method first extracts part-wise base geometric feature and structural feature. 
   ii) Then \textit{PartVAE} encodes base geometric feature to latent space. We then use a part geometry attention mechanism that learns the contribution of each geometric part. iii) Meanwhile, shape structure is \YL{analyzed} and extracted as structural features. iv) After concatenation, the global geometric feature and structural feature are weighted by a geometry structure attention mechanism. v) The weighted geometric and structural feature are concatenated and then encoded by a \textit{GlobalVAE}, whose latent vector is the high-level shape descriptor of the input object.}
   \label{fig:method_pipeline}
\end{figure*}

\subsection{Rotation-invariant Representations}

Rotation invariance is an important attribute of shape descriptors in real-world applications of 3D shape retrieval. Although studies have been conducted by many researchers, this problem is still insufficiently explored. A common technique in this field is discrete feature aggregation.
For instance, \YL{the method \cite{furuya2016deep} by} Furuya et al. extracted shape descriptors from \YL{an} oriented point set by aggregating processed local 3D rotation-invariant features.
Similarly, Qi et al.~\cite{qi2016volumetric} sampled multi-orientation 3D input and aggregated them in a multi-orientation volumetric CNN. Some authors have also suggested using multiple images taken from different viewpoints in their deep models.
Kanezaki et al.~\cite{kanezaki2018rotationnet} mainly focused on how to aggregate predictions from multiple views and take a single image as input for its prediction.
SeqViews2SeqLabels~\cite{han2018seqviews2seqlabels} aggregated the sequential views using an encoder-decoder \YL{Recurrent Neural Network (RNN)} structure with attention. The rotation invariance has been addressed only to a very limited extent because the above methods are not able to deal with arbitrary rotations.

Some recent \YL{research works} suggested incorporating equivariance directly into \YL{the} network architecture, because the desired equivalence of transformation can be achieved through constraining the filter structure.
Thomas et al.~\cite{thomas2018tensor} introduced tensor fields to keep translational and rotational equivariance.
Worral et al.~\cite{worrall2017harmonic} used circle harmonics to \YL{achieve} both translational and 2D rotational equivariance \YL{in} CNNs.
Zhang et al.~\cite{zhang2019quaternion} proposed to represent data by a set of 3D rotations and defined quaternion product units \YL{to} operate on them.

Another way to achieve equivalence is coordinate transformation. Henriques et al.~\cite{henriques2017warped} fixed a sampling grid according to Abelian symmetry.
Also, equivariant filter orbit \YL{was} the main focus of many recent works. Cohen et al.~\cite{cohen2016group} \YL{proposed} group convolution networks (G-CNNs) with the square rotation group. They provided the evidence for the rotational equivariance of group-convolutions.
Worrall et al.~\cite{worrall2018cubenet} proposed CubeNet using Klein’s \YL{four-group} on 3D voxelized data, which learns interpretable transformations with encoder-decoder networks. They \YL{use} a 3D rotation equivariant CNN for voxel representations with linear equivariance to both translation and right angle rotation. Some previous \YL{research works} have applied functions on the icosahedron and their convolutions to achieve equivariance on the cyclic group~\cite{cohen2019gauge} and the icosahedral group~\cite{esteves2019equivariant}.
Esteves et al.~\cite{esteves2018learning} and Cohen et al.~\cite{cohen2018spherical} focused on the infinite group SO(3), and used the spherical harmonic transform for the exact implementation of the spherical convolution or correlation. They proposed to first project 3D objects \YL{onto} a unit sphere and then use spherical convolutional networks to achieve global 3D rotation equivariance. Esteves et al.~\cite{esteves2018learning} also defined several SO(3) equivariant operations on spheres to process 3D data, which can achieve better invariance and generalizes well to unseen rotations. \YL{The} question remains open that how the invariance preservation mechanism can be utilized to learn a shape descriptor for fine-grained shape retrieval.

There are also several recent studies focusing on rotation invariant \YL{representations} on point clouds, which \YL{learn} an initial rotation to a canonical pose.
Qi et al.~\cite{qi2017pointnet} adopted an auxiliary alignment network to make model robust to affine transformations by predicting and applying such transformations to input points and features, which was then further improved to handle the variations in point density by \cite{qi2017pointnet++}.
Deng et al.~\cite{deng2018ppfnet} proposed ordering-free point pair features and a deep architecture based on PointNet to encode coordinates to transform-invariant features. But \YL{these methods} cannot be directly applied \YL{to} mesh models.

\section{RISA-Net}
\label{sec:method}
Our method is inspired by the recent progress in the latent vector learning and transformation-invariant feature extraction. To extract 3D shape descriptors with rich geometric details that are robust to rigid \YL{transformation}, we \YL{propose} to extract \YL{part-wise} mesh-based features: edge \YL{lengths} and dihedral \YL{angles}. These features preserve rigid \YL{transformation invariant} and scale-sensitive geometric details, which enable shape reconstruction from these features~\cite{frohlich2011example}\YL{, showing that complete information is retained}. \YL{Although} these geometric features are descriptive, \YL{their} high-dimensionality means it would be inefficient to \YL{use them directly} as shape descriptors. Moreover, these features only describe low-level feature of edges. It thus lacks information of the global semantic structure of the 3D shape. To address these issues, we adopt a set of variational autoencoders (VAEs) with attention mechanism to extract compact features from the base geometric features, which not only retains the translation and rotation invariance of detailed geometric features, but also balances structure and geometric information to a high-level succinct feature for retrieval tasks. In the following subsections, we first symbolize the elements of RISA-Net, and then introduce the components of the network.

\subsection{Overview}
Fig.~\ref{fig:method_pipeline} illustrates the network architecture of RISA-Net. Given a 3D shape $M_{i}$, the input of the network is its semantically segmented parts $\left\{{m_{i}}^{p}, \forall p \in\left\{1, 2, ..., P\right\} \right\}$, where $P$ is the number of parts. We extract its base geometric feature $ {f_{i}}^{p} $ from each part ${m_{i}}^{p}$. Then we use a set of \textit{partVAE}s (part-wise variational autoencoders) to encode a geometric feature set $\left\{{f_{i}}^{p}, \forall p \in\left[1, P\right] \right\} $. Each partVAE encodes the geometric feature of the corresponding part edge-wisely to a latent vector ${z_{i}}^{p}$. Furthermore, we adopt a part-geometry (\textit{Part-Geo}) attention mechanism to weight the importance of each semantic part to amplify the effect of important parts by multiplying the latent vector of each part by the attention weight ${\alpha^{p}}_{i}$. The weighted latent vector set $\left\{{z'_{i}}^{p}, \forall p \in \left\{1, 2, ..., P\right\}  \right\}$ encodes the geometric information and the contribution of each part to shape discrimination. All the vectors are then concatenated to form a global geometric feature vector $gv_{i}$, representing the geometric feature of the whole shape. Similarly, we extract the global structural feature $sv_{i}$ that is invariant to rigid \YL{transformation} through part-based structure analysis. \YL{As the contributions of geometric and structural features to fine-grained retrieval vary in different cases,} we further learn the importance of geometric and structural features respectively through geometry-structure (\textit{Geo-Struct}) attention mechanism. \YL{In particular}, we multiply the global geometric feature $gv_{i}$ and the global structural feature $sv_{i}$ by the learned geometry weight ${w_{i}}^{g}$ and structure weight ${w_{i}}^{s}$ respectively. Finally, the weighted geometric and structure features are concatenated to get the initial global feature vector $fv_{i}$, which is then interpreted as a low-dimensional latent vector $zv_{i}$ by the global feature variational autoencoder (\textit{GlobalVAE}). $zv_{i}$ will be used as the shape descriptor of the input 3D shape for the fine-grained retrieval task. Additionally, we append triplet loss term to the original VAE loss to improve the distribution of shape features in the latent space and train our attention mechanisms \YL{in an end-to-end manner}.

\subsection{Geometric Feature Representation}
In our observation, globally similar objects could share similar features in some semantic parts, but differ drastically in other parts, since 3D objects are often designed and assembled using different parts to satisfy various desired functions. Therefore, compared with learning features at a low level of granularity for an integrated 3D shape, learning them part-wisely is more effective. In real world applications, 3D objects may be randomly placed, thus are not always strictly aligned or zero-centred to the world coordinate system. Therefore, shape descriptors of 3D objects should be invariant to possible rigid \YL{transformations}. In case of scaling transformation, the feature needs to be capable of describing the relative size of parts for shape discrimination. For example, comparing a coffee table with a dinning table that has the same panel size, the coffee table has shorter legs. Given the above observations, the part-wise geometric feature we \YL{extract} should be invariant to rotation and translation but sensitive to scaling. Finally, we want the extracted features to contain as \YL{much} geometric \YL{detail} as possible, so that the whole 3D shape can be reconstructed from them. Therefore, we represent shapes by edge lengths and dihedral angles, which are reconstructive, scale sensitive and robust to rigid transformations~\cite{frohlich2011example}.

In particular, the base geometric feature is defined on a set of 3D models that all have the same semantic parts with label $p \in \{1, 2, \cdots, P\}$ and $E$ edges with the same connectivity among them, where $P$ and $E$ are the number of Part Semantics and edges respectively. We denote the parts with the same label $p$ of all the models by the set $\left\{m_{i}^{p}, \forall i \in\left[1, N\right] \right\}$, where $N$ is the total number of models. The \YL{same} topological structure of all the shapes can be utilized to establish part-level correspondences. %
In our implementation, we use a \rao{watertight} unit cube mesh with $3075$ vertices as the reference model, and perform non-rigid coarse-to-fine registration~\cite{zollhofer2014real} on the above shape part set, ensuring that all the shapes have the same connectivity as the reference. Each shape $m_{i}^{p}$ could be described by its edge lengths and dihedral angles:
\begin{equation}
    f_{i}^{p} = \left\{ L_{i}^{p},\, \Theta_{i}^{p} |\, L_{i}^{p}\in \mathbb{R}_{+}^{E},\, \Theta_{i}^{p}\in{[0, 2\pi)}^{E} \right\}
\end{equation}
where $L_{i}^{p}\in \mathbb{R}_{+}^{E} $ \YL{contains all $E$ edge lengths}, and $\Theta_{i}^{p}\in{[0, 2\pi)}^{E}$ \YL{includes dihedral angles of all edges.}

\begin{figure*}[t]
\includegraphics[width=1.0\linewidth]{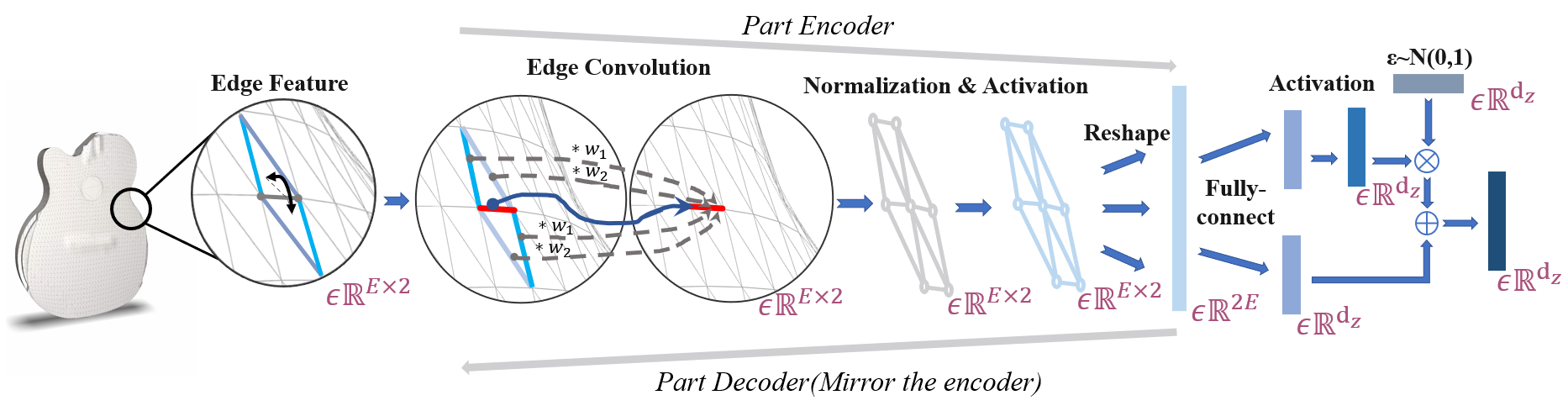}
\DeclareGraphicsExtensions.
\caption{The structure of one partVAE. PartVAE encodes edge features through convolutional operations on edges and their adjacent edges. $E$ denotes the number of edges of a mesh model. $d_z$ denotes the dimension of the latent vector of $\textit{PartVAE}$}.
\label{fig:method_partVAE}
\end{figure*}

The base geometric feature is not ideal for retrieval tasks because \YL{of} its high dimensionality and redundancy. Therefore, after basic geometric information extraction, feature compression is a necessary step. We propose to use \textit{partVAEs} (part-wise variational autoencoders) to extract high-level \YL{features} from the base geometric features. Due to its reconstructive property, the high-level latent vector $z_{i}^{p}$ learned by \textit{partVAEs} will not only maintain the translation and rotation invariance, but also preserve necessary detailed geometric features for object retrieval.

Fig.~\ref{fig:method_partVAE} \YL{illustrates} the structure of one \textit{partVAE}. The input of a \textit{partVAE} is \YL{an} $E\times2$ dimensional feature vector, $f_{i}^{p}$, containing above basic geometry features. In $f_i^p$, it concatenates the rows of an $E\times2$ matrix, where each row \YL{corresponds to} an edge, and \YL{the} two columns represent edge lengths and dihedral angles between two adjacent faces. The encoder of \textit{partVAE} learns the posterior distribution between the input data $f_{i}^{p}$ and the latent vector $z_{i}^{p}$. As the encoder learns local geometric features, convolutional operations on undirected edges are required. Thus, we adapt $MeshConv$ operation~\cite{hanocka2019meshcnn} to encode our reconstructive base geometric feature. In particular, the input is filtered by the first three convolutional layers, where the convolutional operation on the $i^{\rm th}$ edge $e_{i}$ is defined as:
\begin{equation}
    y_{i} = W_{e} \ast x_{i} + W_{n_{e, 1}} \ast \frac{\sum_{j \in N_{e_{i, 1}}}x_{j}}{|N_{e_{i, 1}}|} + W_{n_{e, 2}} \ast \frac{\sum_{j \in N_{e_{i, 2}}}x_{j}}{|N_{e_{i, 2}}|}  + b_{e}
\end{equation}
where $x_{i} \in \mathbb{R}^2$ is the feature \YL{(edge length and dihedral angle)} of edge $e_{i}$. %
\YL{Edges are treated as undirectional, and each mesh part is registered from a unit cube, which is a closed, manifold mesh. Therefore,  $e_i$ is  adjacent to two faces. Within each adjacent face of $e_i$, in the counter-clockwise order, we refer to the edge immediately after $e_i$ as the first adjacent edge, and the edge immediately after the first adjacent edge as the second adjacent edge. Denote by $N_{e_{i, 1}}$ and $N_{e_{i, 2}}$ the sets of first and second adjacent edges of $e_i$, respectively.  In our case, as each edge has 4 neighboring edges, the numbers of elements in each set $|N_{e_{i, 1}}| =  |N_{e_{i, 2}}| = 2$.}
$W_{e}, W_{n_{e, 1}}, W_{n_{e, 2}}\in\mathbb{R}^{2 \times 2}$ are \YL{learnable} weights of convolutional operations on an edge and its adjacent edges. $b_{e}\in\mathbb{R}^{2}$ is the bias term. Additionally, all convolutional layers are appended with a batch-norm layer and a Leaky-ReLU layer with \YL{the slope for negative input} $\tilde{\alpha}=0.02$.

The output of three consecutive convolutional layers is fed into two fully-connected layers to obtain its mean and variance respectively, where the mean $z_{i}^{p}$ is the latent vector of \textit{partVAE}. After that, the decoder learns to reconstruct $f_{i}^{p}$ from the latent vector $z_{i}^{p}$, the network structure of which is symmetric with the encoder without sharing weights. Each \textit{partVAE} is able to extract high-level latent feature of a semantic part, thus depicts geometric information part-wisely. 
\YL{We assume that the number of parts is fixed for all the objects in the same class. However, not all parts need to be present on a given shape. }
If \YL{an} input model misses some parts, the input for the corresponding \textit{partVAE}s will be zero-matrices, and the latent vectors of these parts are set as zeros.

\subsection{Part Geometry Attention Mechanism}
Each semantic part does not contribute equally in shape representation. For example, when measuring the similarity of two car models, car bodies may be more important than car mirrors. Therefore, we further introduce a \textit{Part-Geo} (part geometry) attention mechanism to \YL{learn to determine} the importance of each part of each shape in fine-grained retrieval.

We define an attention vector $\alpha_{i} = [\alpha_{i}^{1}, \alpha_{i}^{2},...,\alpha_{i}^{P}]$ to denote the importance of each part for object $M_i$. The higher the value of $\alpha_{i}^{p}$, the more discriminative the part $p$ is when recognizing shapes. \YL{Following} \cite{vaswani2017attention}, the attention vector $\alpha_{i}$ is obtained by softmax of the dot-\YL{products} of  key vector $K_{i}$ and query vector $Q_{i}$:
\begin{equation}
    \alpha_{i}=softmax(K_{i}^{T}\cdot Q_{i})
\end{equation}
where $K_{i} = [K^{(1)}_{i}, K^{(2)}_{i},...,K^{(P)}_{i}]$ represents the key feature of the part $i$, which is a linear transformation of its latent vector: $K^{p}_{i}=W_{K}^{p}z^{p}_{i}$. The query vector $Q_{i}$ is the summation of the linear transformation of the latent features of all parts: $Q_{i}=\sum_{p}{W^{p}_{Q}}{z_{i}}^{p}$. Here, ${W^{p}_{K}}, {W^{p}_{Q}}\in\mathbb{R}^{d_{h}\times d_{z}}$, $d_{z}$ is the dimension of the latent vector, and $d_{h}$ is the dimension of the key and query features. The attention vector is jointly trained with other parts of the \YL{neural network}.

Thus, the output of the \textit{Part-Geo} attention mechanism is a set of weighted geometric features, $\{\alpha_{i}^{1}{z_{i}^{1}}, \alpha_{i}^{2}{z_{i}^{2}}, ..., \alpha_{i}^{P} {z_{i}^{P}}\}$, \YL{which are concatenated and reshaped} to a vector $gv_{i} \in \mathbb{R}^{P \times d_{z}}$, representing the global geometric information of the object.

\subsection{Structural Information Representation}
Despite the importance of structural information in shape representation, it has been \YL{neglected by existing methods for} shape retrieval. Visually similar shapes often share similar structure. Objects that have parts with similar geometric features can be distinguishable when their structures are highly different. Therefore, we incorporate the global structure of an object as part of our representation.

We represent the structural information by the spatial relationships among the semantic parts. Same as geometric features, the proposed structural features are also robust to rigid \YL{transformation}. For each class of 3D objects, we first define one semantic part as the body part that all models must contain. If there are more than one common semantic parts in all models, we select the part with the largest average volume. We observe that the existence of non-body parts and their relative positions to the body part are important for shape discrimination, which can be used to interpret structural information of shapes. As shown in Fig.~\ref{fig:method_pipeline}, we describe the structure of objects by an 11-dimensional vector $sv_{i}$, \YL{defined as follows:}
\begin{itemize}
    \item $sv_1\in\{0, 1\}$ denotes whether the part exists in the input 3D shape.
    \item $sv_2\in \mathbb{R}$ denotes the distance from the center of the current part to the center of the body part.
    \item $[sv_3, sv_4, sv_5]\in [-1,1]^{3}$ denotes the cosine of the angles between the first \YL{principal} component of the current part and the first three \YL{principal} components of the body part respectively.
    \item $[sv_6, sv_7, sv_8]\in [-1,1]^{3}$ denotes the cosine of the angles between the second \YL{principal} component of the current part and the first three \YL{principal} component of the body part respectively.
    \item $[sv_9, sv_{10}, sv_{11}]\in\mathbb{R}^{3}$ denotes \YL{the} unit direction from the center of \YL{the} body part to the center of the current part. 
\end{itemize}

\begin{figure}[t]
\includegraphics[width=3.4in]{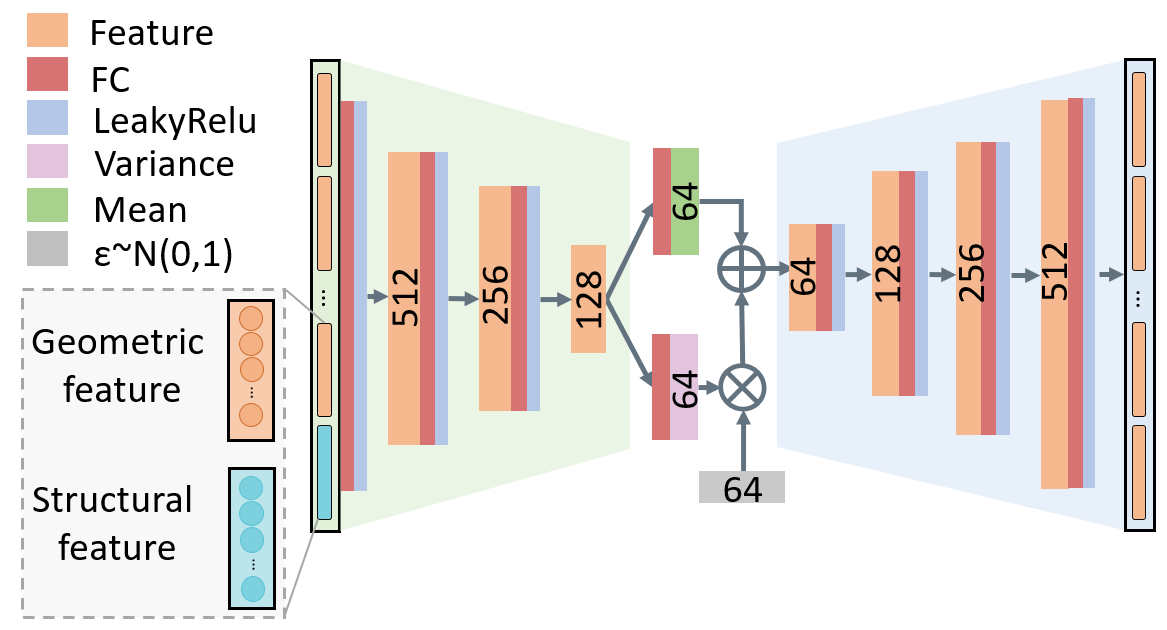}
\DeclareGraphicsExtensions.
\caption{Network architecture of globalVAE. GlobalVAE encodes part geometric features and global structural feature through three consecutive fully-connected layers. The latent vector is the shape descriptor of input shape.}
\label{fig:method_globalVAE}
\end{figure}

\subsection{Geometry-Structure Attention Mechanism}
Although the objects of the same class all share similar structures, objects belonging to different classes have different diversity in the composition. For example, all guitars share the same structure, but chairs have diverse compositions. Thus geometric information and structural information could have different contribution when discriminating objects belonging to different classes.
Therefore, the extracted compact global geometric and structural features $gv_{i}$ and $sv_{i}$ need to be re-weighted to balance their contributions to the final shape representation. 
\YL{To achieve this,} we introduce a \textit{Geo-Struct} (geometry and structure) attention mechanism to learn to balance the importance of structure and geometric information in shape representation.
We define a score vector $w_{i} = [w^{g}_{i}, w^{s}_{i}]\in [0, 1]^{2}$ for model $M_i$, representing the weights of geometric and structural information. The score vector is learned through two fully-connected \YL{sub-}networks:
\begin{equation}
    w_{i}=softmax([F(gv_i), \, G(sv_i))])
\end{equation}
where $F: \mathbb{R}^{d_z}\to [0, 1]$ and $G: \mathbb{R}^{11}\to [0, 1]$. %
In implementation, $F(\cdot)$ contains two fully-connected layers, and the dimension of output vectors of the two layers are $32$ and $1$ respectively. $G(\cdot)$ contains two fully-connected layers as well, where the dimension of output vectors are $16$ and $1$ respectively.
The global geometric feature and structural features are multiplied by \YL{their corresponding weight scores} respectively and then concatenated to form the global feature: $fv_i\in \mathbb{R}^{P \times(d_{z} + 11)}$, which contains weighted geometric and structural information of the object.

\subsection{Global Feature Encoding}
To encode both geometric and structural feature into one latent space, we further use a  \textit{GlobalVAE} (global feature variational autoencoder) to encode global geometric and structural information into a reconstructive compact representation. 
We use the architecture of globalVAE illustrated in Fig.~\ref{fig:method_globalVAE}. The \textit{GlobalVAE} \YL{comprises} three fully-connected layers, \YL{each} appended with a leaky Relu layer with \YL{the slope of negative input $\tilde{\alpha}=0.02$}. 
The structure of \textit{GlobalVAE} ensures that its latent vector $zv_{i}$ contains the geometric information of all parts as well as the global structural information.

\subsection{Losses}
To optimize the network parameters of our model, we adopt three loss-terms that enable a distriminative latent space for fine-grained shape retrieval.

\textbf{VAE Losses.} 
Our optimization objective function includes a Kullback–Leibler (KL) divergence term and a reconstruction term for each \textit{partVAE} and the \textit{GlobalVAE}. The KL \YL{divergence} term regularizes the latent space, while the reconstruction term ensures that the input features can be explained by our autoencoders.
Therefore, the loss function for \YL{all the} partVAEs includes the KL divergence terms and the terms \YL{measuring the} differences between all the input base geometric \YL{features} and their decoding \YL{results}:

\begin{align}
\begin{split}
L_{VAE}^{part}
&= \frac{1}{P_{i}}\sum_{p=1}^{P_{i}} (f_{i}^{p}-{f'}_{i}^{p})^{2} \\
&+ \gamma \sum_{p=1}^{P_{i}}D_{KL_p}^{p}(q(z_{i}^{p}|{f'}_{i}^{p})\parallel p(z_{i}^{p}))
\end{split}
\end{align}
where $P_{i}$ is the number of the parts of model $M_{i}$, $f_{i}^{p}$ is the base feature of the the $p^{\rm th}$ part, ${f'_{i}}^{p}$ is the reconstructed feature of the $p^{\rm th}$ part. In the second term, $\gamma$ is a weight \YL{that balances both terms}, $p(z_{i}^{p})$ is the prior probability distribution, $q(z_{i}^{p}|{f'}_{i}^{p})$ denotes the posterior probability, and $D_{KL_{part}}^{i}$ denotes the KL divergence of the $p^{\rm th}$ partVAE. $\gamma$ is a constant, which is set as $1\times10^{5}$ \YL{in our experiments}. We define the loss for the \textit{GlobalVAE} in a similar way.%

\textbf{Triplet Losses.} Using above losses for VAEs, the distribution of the latent vectors is able to cluster the models of the repository in the feature space to some extent. However, it \YL{can} be further optimized by minimizing the distance between the features of similar shapes and enforcing a margin between dissimilar shapes. Besides, %
we use a triplet loss~\cite{schroff2015facenet} to optimize the final feature distribution in the latent space, which also helps the attention mechanisms to find the distinguished parts and balance the importance of structure information.

For the globalVAE, we define the term as:
\begin{equation}
L_{triplet}^{global} =\sum_{i}[ D(zv_{i}^{a}, zv_{i}^{p}) - D(zv_{i}^{a}, zv_{i}^{n}) + \eta]_{+}
\end{equation}
where $zv_{i}^{a}$, $zv_{i}^{p}$ and $zv_{i}^{n}$ are the latent features of 
\YL{an anchor model (i.e., a chosen model in the training iteration), a positive model (i.e., a model of the same sub-class as the anchor model) and a negative model (i.e., a model of a different sub-class from the anchor model). } \rao{$D(\cdot,\cdot)$ is a measure of distance between two vectors in the latent space.}
\YL{We use the Euclidean distance $D(v_1, v_2) = \|v_1 - v_2\|^2_2$ in our experiments.}
\YL{$\eta$}
is the threshold of the margin between the distances from the reference model to the similar and dissimilar models. \rao{In implementation, the Euclidean distance $D(\cdot,\cdot)$ between features are normalized to $[0, 1]$ in each batch. }\rao{We set $\eta$ to 0.3 in the following experiments.}

For the set of \textit{partVAE}s, we define a triplet loss term as:
\begin{equation}
L_{triplet}^{part} =\sum_{i}[ D({gv}_{i}^{a},  {gv}_{i}^{p}) - D({gv}_{i}^{a}, {gv}_{i}^{n}) + \eta]_{+}
\end{equation}
We use the term to refine the distribution of the global geometric \YL{feature} $gv_{i}$ for the entire set of partVAEs.

\textbf{Overall Loss.} The overall loss of RISA-Net is:
\begin{equation}
L =L_{VAE}^{part} +\lambda_{1} L_{VAE}^{global} +\lambda_{2} L_{triplet}^{part} + \lambda_{3} L_{triplet}^{global}
\end{equation}
where $\lambda_{1}$, $\lambda_{2}$ and $\lambda_{3}$ are hyper-parameters to balance the weights of different loss terms, which are defaultly set as $1\times{10}^3$, $1\times{10}^2$, and $1\times{10}^2$ in our all experiments.

\subsection{Model Training and Shape Retrieval}

We feed 3D shapes of all the sub-classes of the same class to RISA-Net to learn to encode base features into a latent feature for that class. The \textit{partVAE} set with \textit{Part-Geo} attention mechanism first learns a high-level geometric feature set from base geometric features, using the reconstruction loss, KL losses and triplet losses. \YL{At} the same time, the high-level geometric feature set is balanced with structural feature and then fed to the \textit{GlobalVAE} to learn to generate the final latent feature vector, where we minimize the same three types of loss terms during training. In addition, we use the objects of the same sub-class as similar shapes \YL{and objects of different sub-classes as dissimilar shapes} when minimizing the triplet losses.

For an input 3D object $M_i$, we use the latent vector of the \textit{GlobalVAE} $gv_{i}$ as its shape descriptor. For each query shape, we rank the shapes in the repository according to the Euclidean distance between their shape descriptors. Note that the distance metric used in retrieval is the same as in the triplet loss, which is \YL{the Euclidean} distance.

\section{Experimental Results}
\label{sec:experiments}
In this section, we describe our dataset, implementation details and the experiments to validate the effectiveness of RISA-Net. We build a RISA-dataset with 3D shapes labeled \YL{at the sub-class} level to support experiments of fine-\YL{grained} retrieval and provide both quantitative and qualitative comparisons between RISA-Net and other state-of-the-art methods. We also show shape retrieval results on other publicly available datasets to demonstrate the generalizability of the shape descriptors extracted by our RISA-Net. Finally, we conduct ablation studies to verify the \YL{necessity and} complementarity of all the components of our model.

\subsection{Dataset}
\label{sec:dataset}
Retrieving a model against shapes within the same class but belonging to different sub-classes is a typical fine-grained shape retrieval task, where the repository contains globally similar shapes \YL{that} differ in some details. Most of the existing large-scale 3D object datasets are annotated with only \YL{class level} labels, which is not suitable for fine-grained retrieval task. For example, ModelNet~\cite{wu20153d} contains 662 object categories but only a few of them are given sub-class labels. Part of ShapeNet models\cite{chang2015shapenet} also have intra-class semantic labels, but the \YL{labeling} precision and amount are not sufficient to test intra-class retrieval methods. \YL{Although} FRGC v2 dataset~\cite{phillips2005overview} is labeled with an intra-class manner, the dataset only focuses on facial recognition rather than common objects retrieval. Therefore, we build a new dataset with \YL{sub-class level annotations, designed for training, evaluating and comparing fine-grained shape retrieval methods, which is used to demonstrate the effectiveness of our latent descriptor for fine-grained shape retrieval.}
It could be used to support and evaluate future \YL{research} on fine-grained 3D shape classification and retrieval task.
The shapes used in our fine-grained 3D object retrieval dataset is a subset of SDM-NET data\cite{gao2019sdm}, containing 8,906 3D models from six object categories. The six categories are knife, guitar, car, plane, chair and table, which are further grouped into 175 sub-classes. Each sub-class of models is annotated with semantic labels, which are defined by their distinguishable features compared with other sub-classes, such as functionality, product %
\YL{model number}
and style. Take the category of 
guitar as an example, objects are further categorized into twelve sub-classes including double-neck guitar, acoustic, cutaway, Flying V, Gibson Explore, Gibson Les Paul, etc. These semantic labels are assigned according to their style and standard model %
number.

All of the intra-class labels were annotated manually. For clarification, the original annotation from ShapeNet is used as a reference for our annotators. Additionally, some unrealistic shapes that are hard to be categorized were discarded, because keeping them would confuse the retrieval method by using ambiguous sub-class labels, leading to inaccurate quantitative analysis. Please refer to our supplementary materials for more detailed information of the dataset.

\subsection{Implementation Details}
\label{sec:Implementation}

\subsubsection{Dataset Preparation}
We evaluated the performance of RISA-Net in the fine-grained 3D object retrieval task on \YL{the RISA-Dataset introduced in the previous subsection}. We tested the performance of RISA-Net on all the six object categories. To demonstrate the robustness of RISA-Net \YL{to} rigid \YL{transformation}, we perturbed all models in the dataset by transforming each model with a random rotation in \textit{SO(3)}. All experiments were conducted on the perturbed dataset.

\subsubsection{Training Details}
The experiments were carried out on a machine with an \YL{Intel} i7-6850 CPU, 128GB RAM and a GTX 1080Ti GPU. We randomly split the dataset into training and testing sets with a ratio of 4:1. The network is trained in \YL{an} end-to-end manner. In optimization, we adopt Adam Optimizer\cite{kingma2014adam} and set the learning rate to constant $1 \times 10^{-5}$. The batch size of each iteration is 8. \rao{The optimization \YL{repeats} until the loss converges.}

\begin{table}[!ht]
\renewcommand{\arraystretch}{1.3}
\caption {Comparison of \YL{average} performance on six categories with different dimensions of latent \YL{vectors}.}
\label{table:set_latent}
\centering
\begin{tabular}{|c|c|c|c|}
\hline
($d_z$, $d_s$)    & (32, 32)    & (32, 64)             & (32, 128)   \\ \hline
mAP(micro, macro) & (0.53,0.44) & (0.49,0.36)          & (0.49,0.39) \\ \hline \hline
($d_z$, $d_s$)    & (64, 32)    & (64, 64)             & (64, 128)   \\ \hline
mAP(micro, macro) & (0.59,0.51) & \textbf{(0.61,0.51)} & (0.22,0.15) \\ \hline \hline
($d_z$, $d_s$)    & (128, 32)   & (128, 64)            & (128, 128)  \\ \hline
mAP(micro, macro) & (0.55,0.47) & (0.57,0.49)          & (0.49,0.38) \\ \hline
\end{tabular}
\end{table}

\subsubsection{Parameter Setting}
We adopt the same network structure and hyper-parameters on all six object categories. \textit{GlobalVAE} comprises 3 fully-connected layers, the dimensions of which are set to 512, 256, and 128 respectively. We evaluated the mean average precision (mAP) of the retrieval results on the six object categories with different dimensions of the latent vectors of \textit{PartVAE} and \textit{GlobalVAE}, and find that 64 dimensional latent vectors are the most empirically effective settings for both of them. The experimental results on different latent vector settings are illustrated in Table~\ref{table:set_latent}, where $d_z$ and $d_s$ denote dimensions of the latent \YL{vectors} of partVAE and globalVAE respectively.

\begin{figure*}[htbp]
    \centering
    \subfigure[Knife]{
        \includegraphics[width=0.31\linewidth]{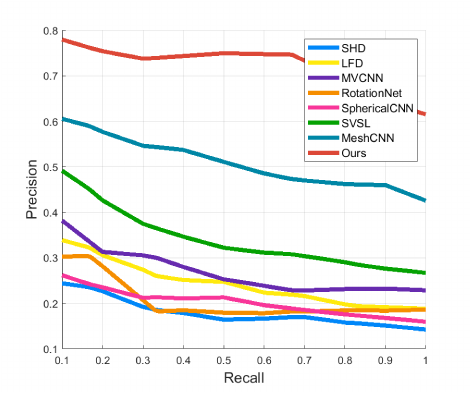}
    }
    \subfigure[Guitar]{
	\includegraphics[width=0.31\linewidth]{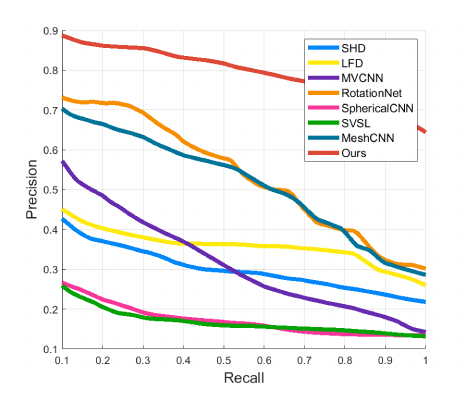}
    }
    \quad
    \subfigure[Car]{
    	\includegraphics[width=0.31\linewidth]{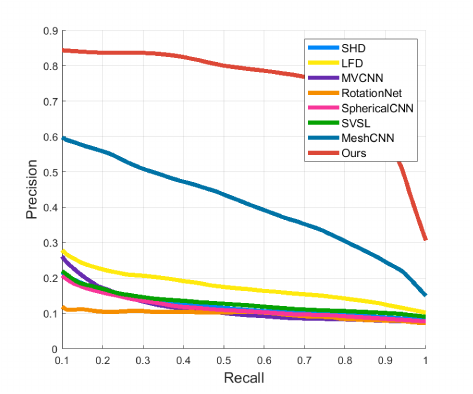}
    }
    \subfigure[Plane]{
	\includegraphics[width=0.31\linewidth]{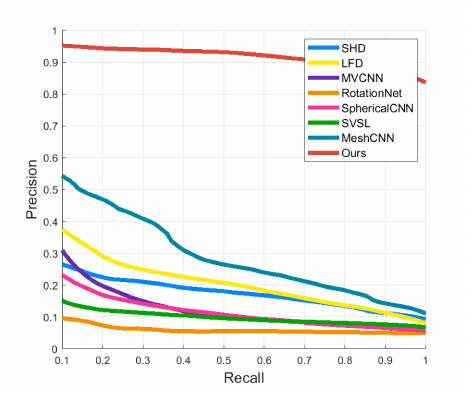}
    }
    \quad
    \subfigure[Chair]{
    	\includegraphics[width=0.31\linewidth]{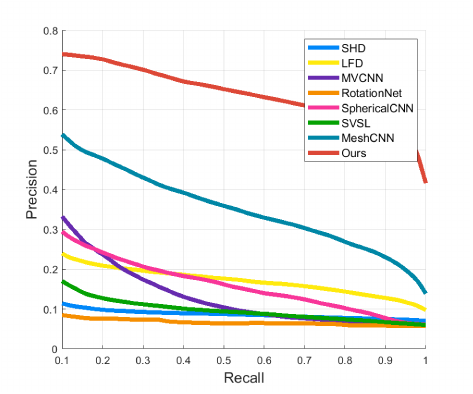}
    }
    \subfigure[Table]{
	\includegraphics[width=0.31\linewidth]{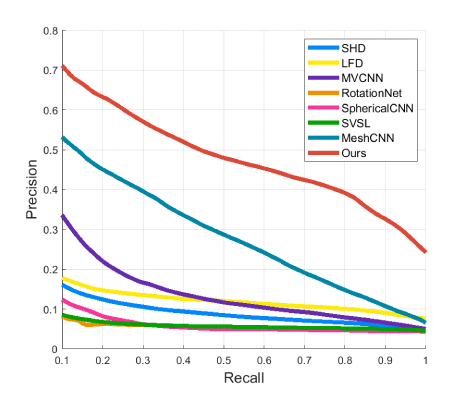}
    }
    \caption{\YL{Comparison of Precision-Recall Curves of different methods} on six object categories.}
    \label{fig:exp_PR_general}
\end{figure*}

\begin{table*}[!t]

\caption{Comparison between RISA-Net and other methods using five metrics on six object categories. \YL{The reported results are based on the test set for fairness.}}
\vspace{.1cm}
\label{table:exp_measure_general}
\centering
\begin{tabular}{|c|c|c|c|c|c|c|c|c|c|c|}
\hline
\multicolumn{1}{|l|}{} & \multicolumn{5}{c|}{micro}  & \multicolumn{5}{c|}{macro}                                                              \\ \hline
 Methods               & NN              & FT              & ST              & NDCG            & mAP             & NN              & FT              & ST              & NDCG            & mAP             \\ \hline
 SHD~\cite{kazhdan2003rotation}               & 0.1620          & 0.1335          & 0.2450          & 0.4358          & 0.1617          & 0.0904          & 0.0748          & 0.1461          & 0.3516          & 0.1107          \\
                           LFD~\cite{chen2003visual}                  & 0.2241          & 0.1827          & 0.3134          & 0.4860          & 0.2143          & 0.1277          & 0.1065          & 0.1929          & 0.4010          & 0.1525          \\
                           MVCNN~\cite{su2015multi}                    & 0.3374          & 0.1680          & 0.2682          & 0.4864          & 0.1884          & 0.2324          & 0.1223          & 0.1874          & 0.4123          & 0.1492          \\
                           RotationNet~\cite{thomas2018tensor}                & 0.1688          & 0.1385          & 0.2301          & 0.4339          & 0.1717          & 0.1211          & 0.0989          & 0.1515          & 0.3634          & 0.1349          \\
                           Spherical~\cite{esteves2018learning}               & 0.1590          & 0.1195          & 0.2068          & 0.4233          & 0.1404          & 0.1002          & 0.0781          & 0.1289          & 0.3551          & 0.1061          \\
                           SVSL~\cite{han2018seqviews2seqlabels}                   & 0.1604          & 0.1318          & 0.2381          & 0.4573          & 0.1541          & 0.0956          & 0.0832          & 0.1422          & 0.3769          & 0.1095          \\
                           MeshCNN~\cite{hanocka2019meshcnn}               & 0.4653          & 0.2979          & 0.4561          & 0.6739          & 0.2938          & 0.3541          & 0.2421          & 0.3815          & 0.5760          & 0.2448          \\
                           Ours                  & \textbf{0.7378} & \textbf{0.5670} & \textbf{0.6892} & \textbf{0.7878} & \textbf{0.6125} & \textbf{0.6003} & \textbf{0.4642} & \textbf{0.5631} & \textbf{0.7065} & \textbf{0.5142} \\ \hline
\end{tabular}
\end{table*}

\subsection{Fine-\YL{Grained} Shape Retrieval}
In the fine-grained object retrieval task, we query a shape among the shapes of the same class in \YL{the} dataset. As mentioned above, all shapes are perturbed by random rotations in the experiment. RISA-Net is trained to extract latent feature vectors for shape representation, which are then used to measure the similarity between the query and all the shapes of the same class. \rao{The similarity between shapes is measured by the Euclidean distance between shape descriptors.}
If a retrieved shape is in the same sub-class with the query shape, we denote it as a successful retrieval. Fig.~\ref{fig:sample_general} shows the top five retrieved results \YL{for} \rao{four} query shapes on chair category. 
Since RISA-Net is able to find shapes with matched geometric details, it performs well in the sub-class retrieval.  Specifically, as shown in the \rao{first} row of
Fig.~\ref{fig:sample_general}, RISA-Net captures that the query object \rao{has grid cotton pad on the chair back}, 
and then \YL{retrieves} shapes with similar features. Meanwhile, the retrieved results with 
lower rankings 
also show the capability of RISA-Net. As shown in the \rao{last row}, the query shape has a slat back and turned legs. 
RISA-Net successfully \YL{retrieves} shapes with matched features, \YL{which} are in the same sub-class \YL{as} the query shape. For more retrieval results, please refer to the supplementary material.

We compare RISA-Net with other alternative approaches, including Spherical Harmonics descriptor (SHD)~\cite{kazhdan2003rotation}, LightField descriptor (LFD)~\cite{chen2003visual}, MVCNN~\cite{su2015multi}, RotationNet~\cite{thomas2018tensor}, spherical CNNs~\cite{esteves2018learning}, SeqViews2SeqLabels (SVSL)~\cite{han2018seqviews2seqlabels} and MeshCNN~\cite{hanocka2019meshcnn}. For both MVCNN and SeqViews2SeqLabels, we render 12 views of the shape, as reported in their best results.
For RotationNet, we use the same camera setting \YL{as \textit{case(ii)}} in their paper, which places cameras on vertices of a dodecahedron encompassing the object. For spherical CNNs, we use in-batch triplet loss for retrieval. For MeshCNN, we use the output of the classification network as \YL{the} shape descriptor, and rank the shape descriptors by their \YL{Euclidean} distance \YL{to the query} to measure shape similarity. 
All the above methods are evaluated on our perturbed dataset.
Fig.~\ref{fig:exp_PR_general} visualizes the comparative results on the test set of RISA-Net and other methods with the PR \YL{(Precision-Recall)} Curves for \YL{the} six object categories. The performance of the learned descriptors by other methods varies among categories. The performances of SHD and LFD are more stable than learning-based methods. The PR curves show that RISA-Net clearly outperforms other methods on all categories in terms of both precision and recall.

\begin{figure*}[t]
    \centering
    \includegraphics[width=7.0in]{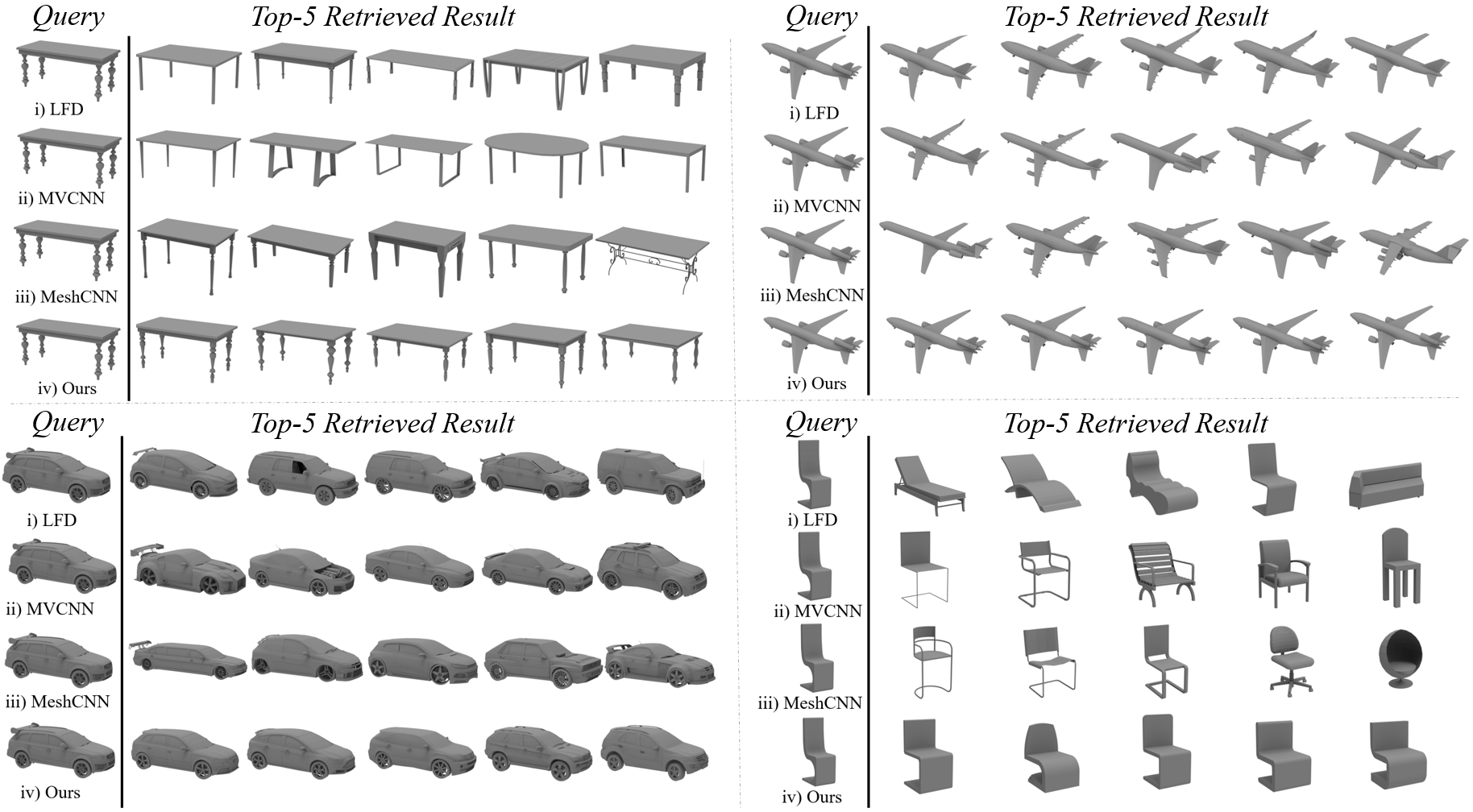}
    \caption{Visual comparison of retrieval results among LFD, MVCNN, MeshCNN and our method. Our method outperforms the other methods on geometric local feature encoding and fine-grained object retrieval.}
    \label{fig:sample_table}
\end{figure*}

Table~\ref{table:exp_measure_general} shows the quantitative comparison between RISA-Net and other methods on six object categories. 
\rao{We conduct quantitative experiment on fine-grained shape retrieval. }\rao{Both query shapes and repository come from the test set to ensure fairness.} The results are evaluated using various statistical metrics: Nearest Neighbor (NN), First Tier (FT), Second Tier (ST), Normalized Discounted Cumulative Gain (NDCG) and mean Averaged Precision (mAP). Considering the imbalance of model numbers of the sub-classes, each measure is calculated through both micro and macro average. The macro averaging computes the metric independently for each sub-class and then \YL{takes} their average as the final overall metric, whereas the micro averaging 
\YL{is a weighted average with the weight for each sub-category proportional to the number of objects in it}.
Table~\ref{table:exp_measure_general} demonstrates the performances of RISA-Net and other methods on the six categories of objects. Noticeably, mesh-based deep models, MeshCNN~\cite{hanocka2019meshcnn} and our method \YL{outperform} other methods on almost all of the metrics. \rao{Apart from these two methods, MVCNN\cite{su2015multi} and LFD\cite{chen2003visual} are better than other methods on all of the metrics.} Overall, RISA-Net achieves the best performance among all alternative methods under all the above metrics.

Fig.~\ref{fig:sample_table} provides examples of top-5 retrieved results by using LFD, MVCNN, MeshCNN and RISA-Net respectively. In the top-left example, the query shape is a traditional rectangular dining table with turned legs. The retrieved models by LFD are with the similar height, surface and leg numbers, which match the coarse-level features of the query shape. MVCNN returned models are with similar height but failed to recognize the rectangular top and turned legs. MeshCNN found rectangular tables with decorated legs, but the shapes of their legs \YL{do not match the query}, and some tables are obviously taller than the query table. 
Only RISA-Net successfully recognized the turned table legs, showing its superiority on encoding fine-grained geometric features. In the top-right example, though all three methods retrieve globally similar planes, only RISA-Net identified the discriminative feature on the tail of the query plane and retrieved shapes of the same sub-class, showing its capability of recognizing discriminative parts. In the bottom-left example, the query shape is a hatch-back car, however, most of LFD's retrieved results are minivans, most of MVCNN's retrieved results are sedans, and only \YL{some} of MeshCNN's retrieved cars are visually similar to the query shapes. Our method not only returned results that are all hatch-backs \YL{rather} than cars from other sub-classes, but also \YL{returned} hatch-backs with wheels similar to the query shape. In the bottom-right example, robustness to rigid \YL{transformation} and the balanced structure/geometric information enabled our method to retrieve chairs of the same sub-class, but other methods mostly failed.

\begin{figure}[htbp]
\centering
\includegraphics[width=3.4in]{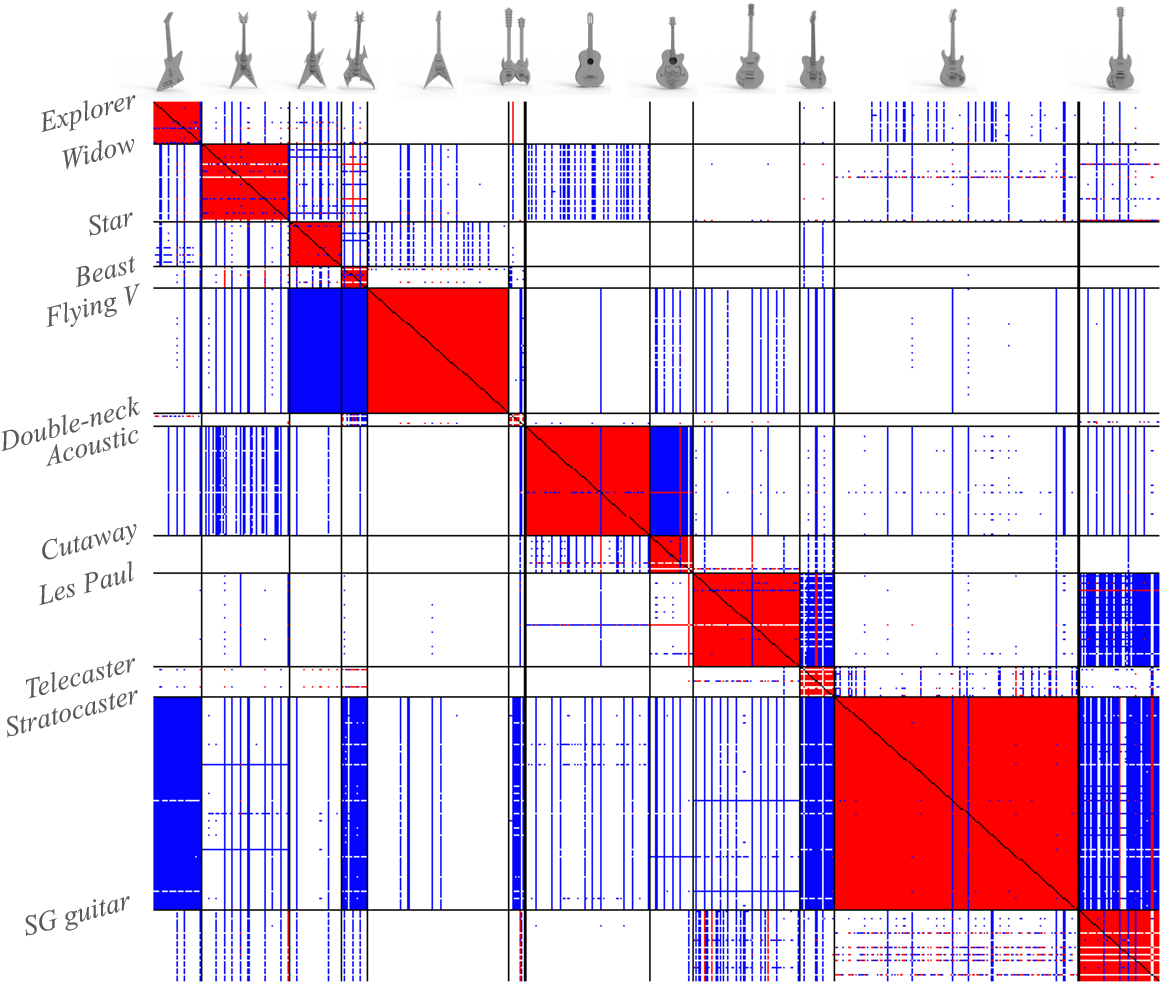}
\DeclareGraphicsExtensions.
\caption{Illustration of \YL{the tier image of our result} for the dataset of guitar. }
\label{fig:exp_tier_guitar}
\end{figure}

In Fig.~\ref{fig:exp_tier_guitar}, we use a tier image \cite{benchmark} to visualize the global retrieval results \YL{of our method}. In a tier image, each row represents a query with model $j$. Pixel $(i, j)$ is filled by black, red, and blue if model $i$ is the nearest neighbor, first tier match, and the second tier match of $j$ respectively. Along each axis, models are grouped by sub-class, and lines are added to separate each sub-class. In each sub-class, red pixels are clustered in blocks along the diagonal, showing that models of the same sub-class are each other's first tier matching results. Moreover, second tier matches of each sub-class tend to congregate together in the same block, implying that RISA-Net learns the similarity between sub-classes. For example, most second tier matches of \textit{Les Paul} are under the sub-class of \textit{Telecaster} and \textit{SG guitar}. \YL{It can be observed} that the three sub-classes of guitar share many local features as shown by the illustrated randomly selected instances. Both \textit{Telecaster} and \textit{SG guitar} are electric guitars that have a thin solid body as \textit{Les Paul}. Also, these guitars have similar round shapes in their lower-half bodies and missing cuts on their upper half bodies.

\begin{figure}[htbp]
    \centering
    \subfigure[plane]{
        \includegraphics[width=3.5in]{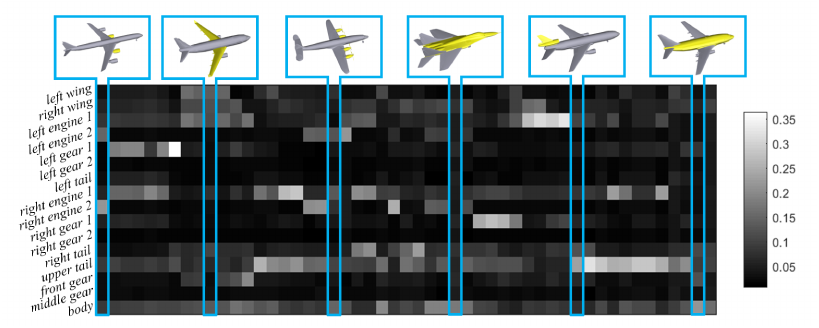}
    }
    \subfigure[chair]{
	    \includegraphics[width=3.5in]{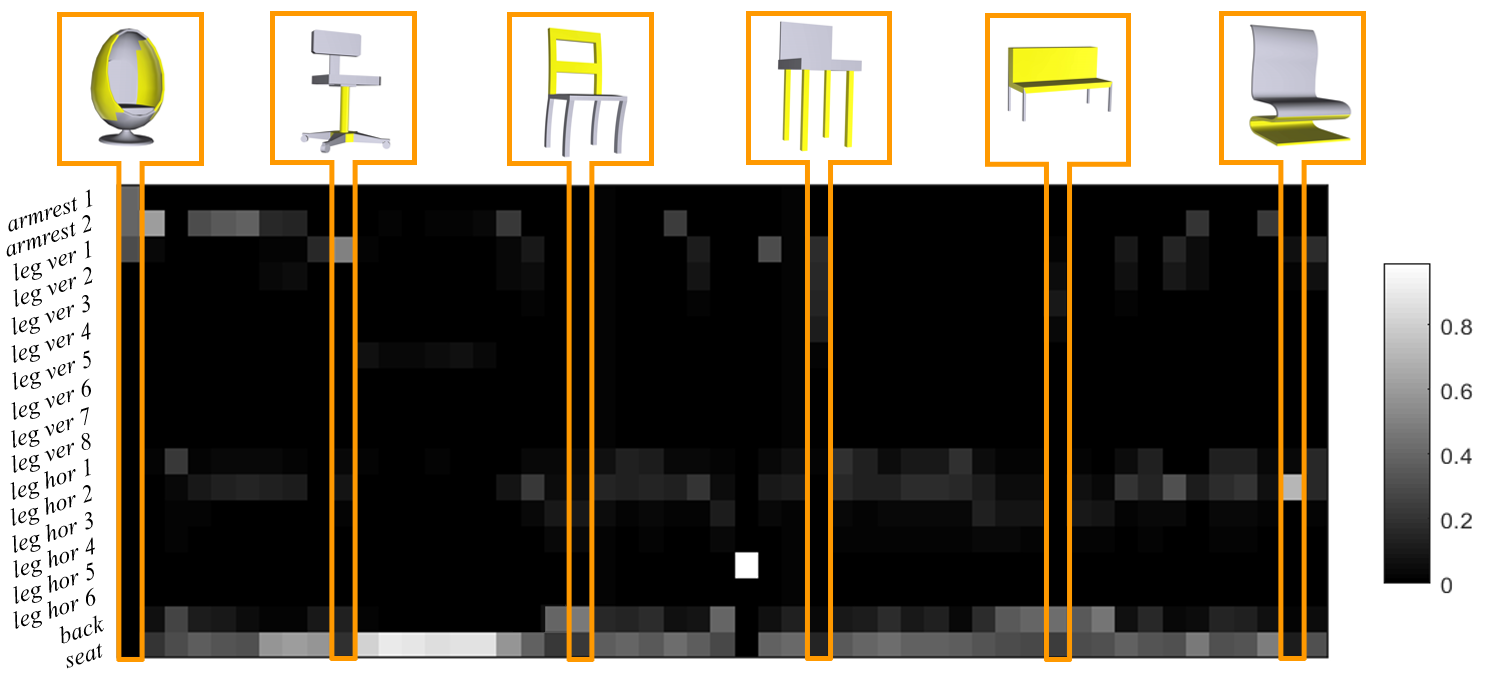}
    }
    \caption{Part geometry attention \YL{visualization on the plane and chair datasets}.}
    \label{fig:exp_attention}
\end{figure}

\subsection{Weighted Features of Parts by Part-Geo Attention}
Our model utilizes the \textit{Part-Geo} attention mechanism to balance the contributions of different parts when discerning a shape among sub-classes. Fig.~\ref{fig:exp_attention} visualizes the learned attention information that demonstrates the importance of each part when learning the final latent descriptor, where we highlight the valid parts using the learned attention weights. Note that the weights of missing parts are set to $0$. Aside from those missing parts, the discriminative parts of the shapes are successfully assigned with relatively high weights, meaning that they contribute more than the other parts to the latent feature learning. More specifically, in Fig.~\ref{fig:exp_attention}(a), the highest weight of each plane appears on the most discriminative part: engine, wing, tail or body. In Fig.~\ref{fig:exp_attention}(b), the weights of existing parts also conform to their discriminativeness. The highest weights are assigned to leg, arm, back and seat respectively. %

\begin{figure}[htbp]
    \centering
    \subfigure[guitar]{
        \includegraphics[width=3.5in]{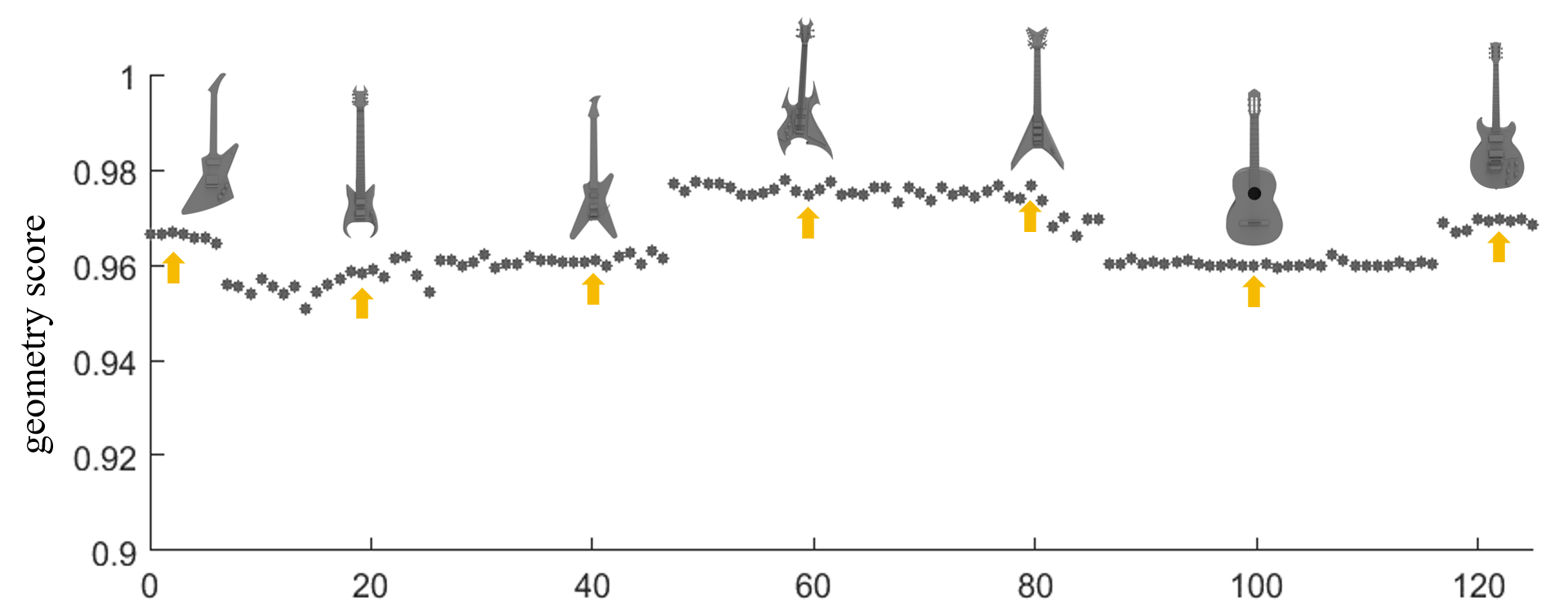}
    }
    \subfigure[chair]{
	    \includegraphics[width=3.5in]{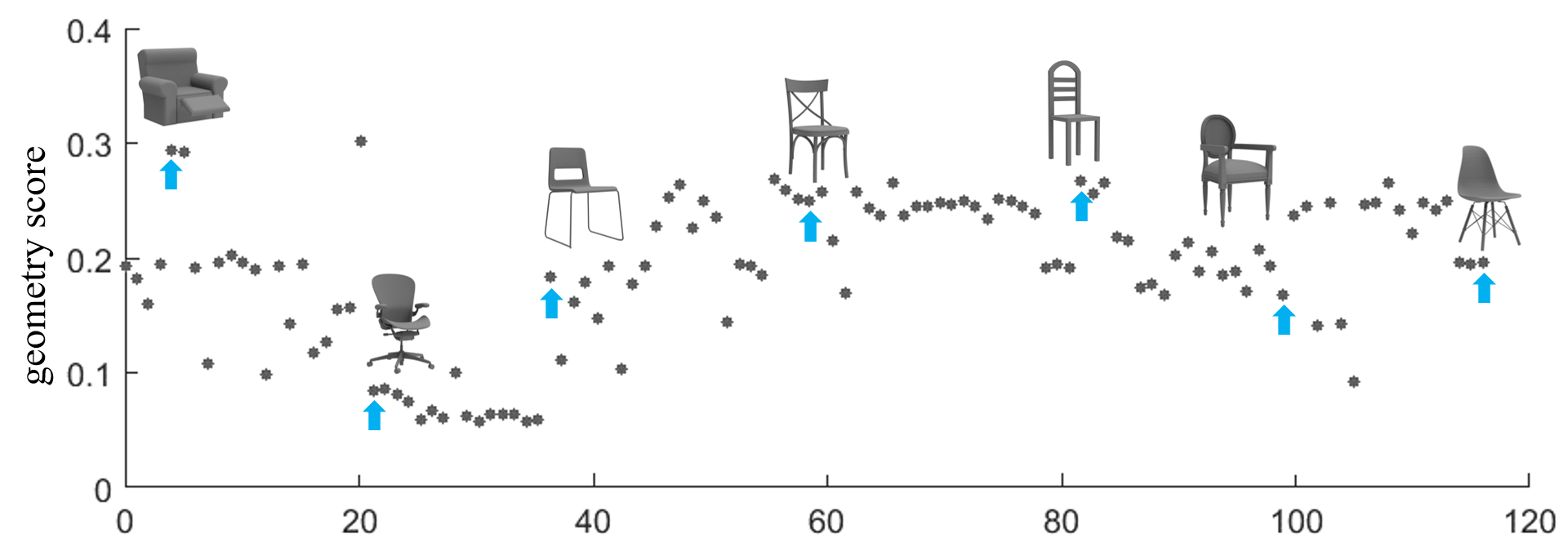}
    }
    \caption{Visualization of geometry-structure attention information on guitar and chair datasets.}
    \label{fig:exp_gs_attention}
\end{figure}

\subsection{Weighted Features by Geo-Struct Attention}
The \textit{Geo-Struct} attention mechanism balances the geometry and structure information in our shape representation. Fig.~\ref{fig:exp_attention} visualizes the learned geometry score $w^g$ in guitar and chair \YL{datasets}. The structure score is calculated by: $w^s = 1-w^g$. In each dataset, we randomly selected some model in the test set. Models are grouped by sub-classes along $x$-axis, and their geometry scores are shown by their $y$-coordinates. Each point represents an object instance, and we draw the 3D shapes of several representative instances for better visualization. Note that shapes of the same sub-class that share similar structure/geometry features tend to have similar structure/geometry scores. Also, the average geometry score of the class of guitar is higher than chair, indicating that distinguishing guitars relies more on geometric information than chairs.

\subsection{Generalizability}

\begin{figure}[b]
\begin{center}
\includegraphics[width=3.0in]{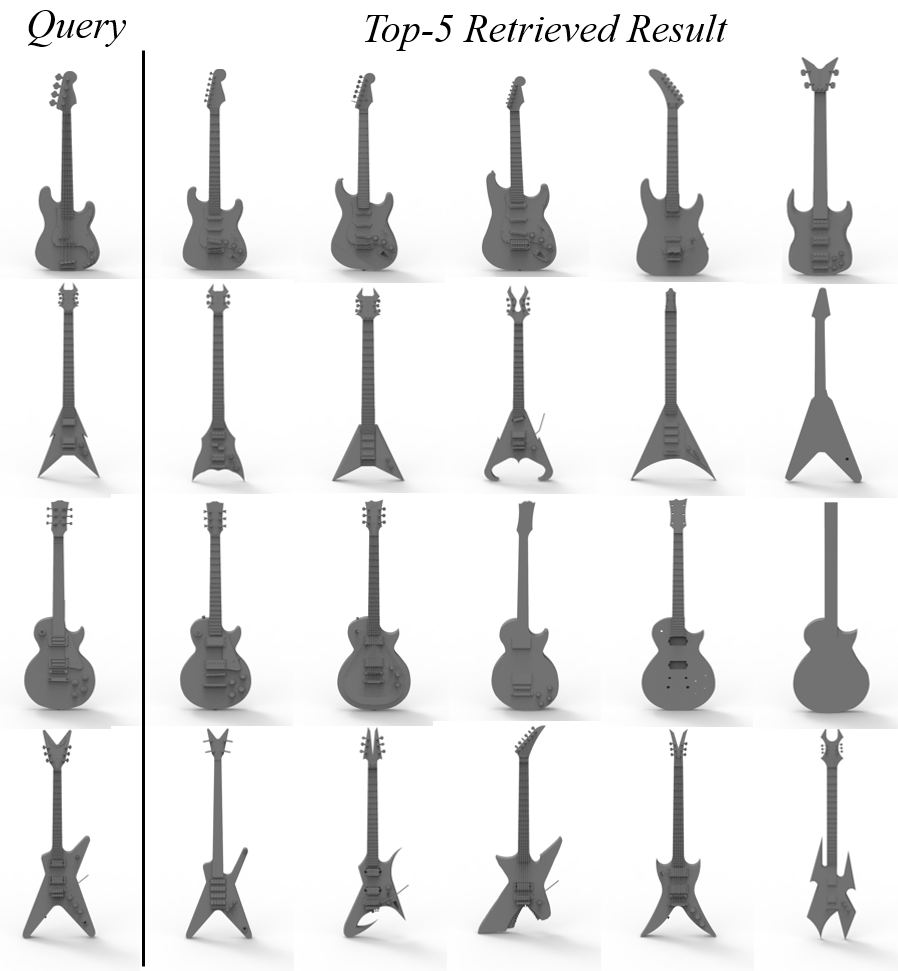}
\end{center}
   \caption{Examples of top-5 retrieval results on the test set of ModelNet guitar dataset. }
\label{fig:exp_ModelNet}
\end{figure}

\textbf{Performance on other public datasets.}
To validate the ability \YL{of RISA-Net} to generalize to other public datasets, we manually segmented the test dataset of the guitar category of ModelNet40, which contains 100 guitar models. All segmented parts are registered from a unit cube with $3075$ vertex in a coarse-to-fine manner~\cite{zollhofer2014real}. Then, using the network trained on our dataset, we feed the manually segmented mesh directly to the trained model without any fine-tuning operation. Some results are illustrated in Fig.~\ref{fig:exp_ModelNet}, which demonstrate that RISA-Net is capable of learning effective features
\YL{that generalize well to new datasets}.

\begin{figure}[htbp]
    \centering
    \subfigure[guitar]{
    	\includegraphics[width=1.65in]{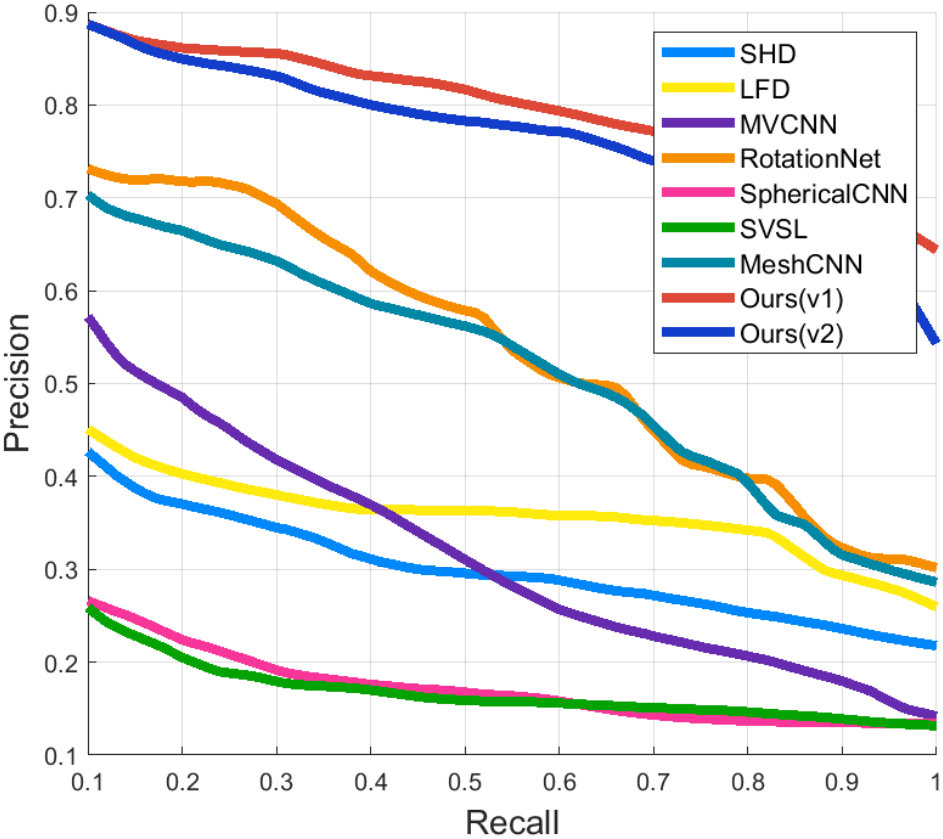}
    }
    \subfigure[plane]{
        \includegraphics[width=1.65in]{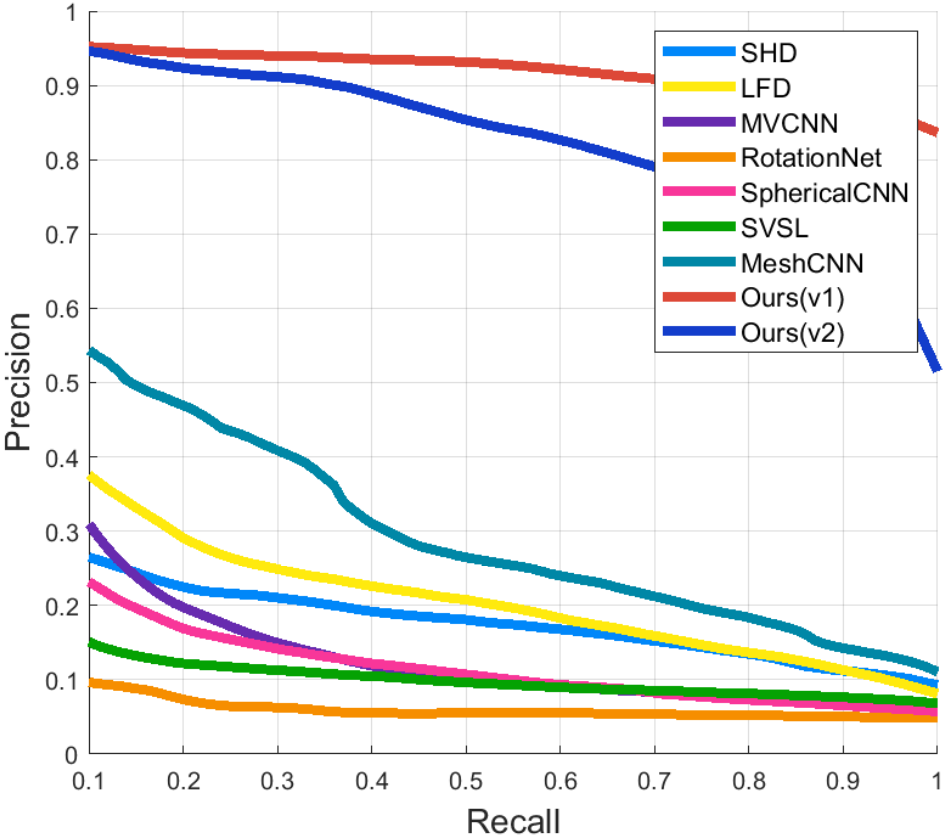}
    }

    \caption{PR \YL{curves} on guitar and plane \YL{datasets} using models without pre-segmentation}
    \label{fig:exp_seg}
\end{figure}

\textbf{Retrieval with fully-automatic part segmentation. }
In this experiment, we investigate whether our approach is sufficiently practical to be combined with existing shape segmentation methods to retrieve a shape without pre-segmentation. Instead of manually segmenting the shapes into parts, we use the following way to automatically obtain the segmentation results. We randomly and uniformly sampled 2048 points from \YL{each watertight}
mesh, and then feed the points to PointNet++ \cite{qi2017pointnet++} for segmentation. The segmentation accuracy (the ratio of the correctly segmented) on guitar and plane datasets are 92.46\% and 89.29\% respectively. With the semantically labeled points from PointNet, we align \YL{watertight} mesh models to the points and assign each triangular mesh with the label of the closest point. Then we use the segmented meshes to train and test our model. Fig.~\ref{fig:exp_seg} shows the PR curves of RISA-Net's performance on the datasets of guitar and plane using models without pre-segmentation (\textit{Ours(v2)}),  \YL{compared} with the pre-segmented models (\textit{Ours(v1)}). We also display the PR curves of other methods for comparison. Note that our approach is able to \YL{tolerate} minor part segmentation errors, indicating its capability to be used in real world shape retrieval applications.

\subsection{Ablation Study}
We now provide \YL{the} results of a detailed ablation study that shows the contribution of each component to the overall performance, including base geometric feature selection, the feature extractor selection and the hierarchical structure.

\begin{figure}[t]
\begin{center}
\includegraphics[width=3.0in]{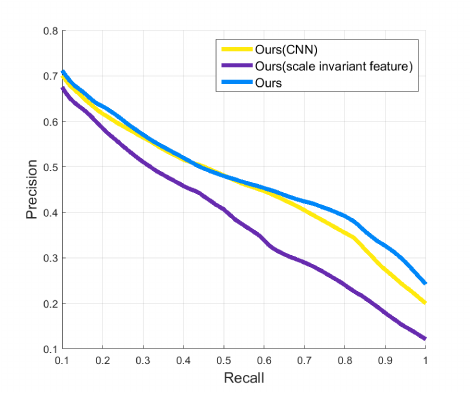}
\end{center}
   \caption{The PR \YL{curves} comparing different base features and geometric feature extractors \YL{on the table dataset}.}
\label{fig:exp_meshcnn_pr}
\end{figure}

\textbf{\YL{Geometric} feature extraction.}
In this subsection, we show the effectiveness of the \YL{adopted} scale-sensitive features and a reconstructive network to describe part geometry. We design a comparison group that replaces our base geometric feature with the 5 dimensional shape feature proposed by Hanocka et al.~\cite{hanocka2019meshcnn} as a scale-insensitive feature, which is denoted as ``Ours(scale sensitive)''. We design another comparison group by replacing \textit{PartVAEs} with the classification network in \cite{hanocka2019meshcnn}, which is denoted as ``Ours(CNN)''. The PR curves of fine-grained retrieval results on the table category with these \YL{three} configurations are demonstrated in Fig.~\ref{fig:exp_meshcnn_pr}. It shows that \YL{the} scale-sensitive base feature is more suitable for our framework, and \YL{the} reconstructive network is better for detailed geometric information learning.

\begin{figure}[htbp]

    \centering
    \subfigure[Comparison between scale-invariant and scale-sensitive feature as base geometric feature]{
        \includegraphics[width=3.5in]{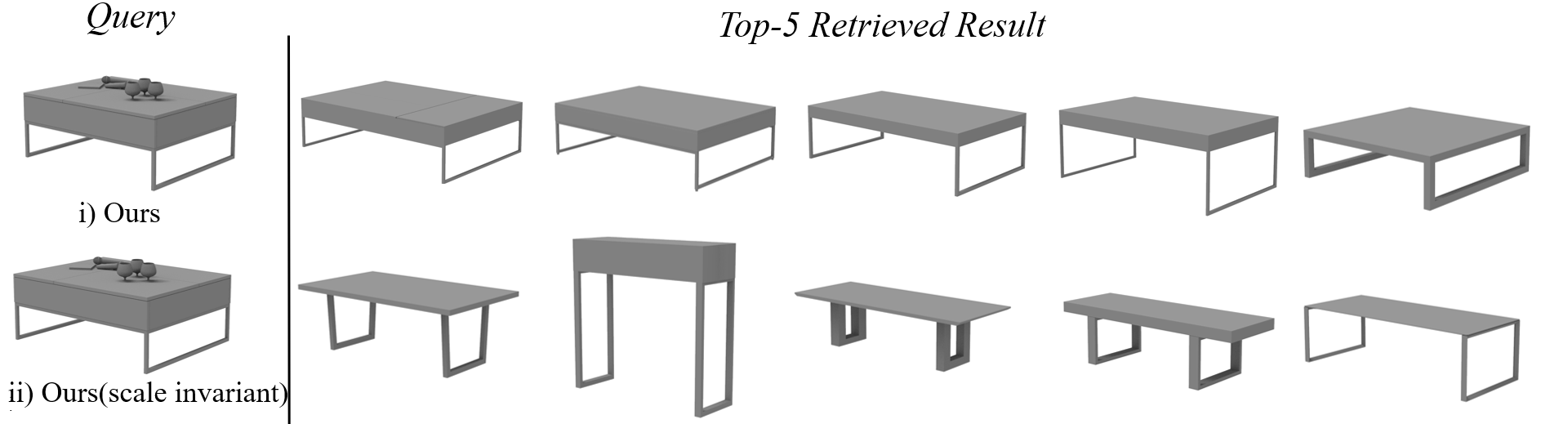}
    }
    \subfigure[Comparison between classification network and reconstructive network as base geometric feature extractor.]{
	    \includegraphics[width=3.5in]{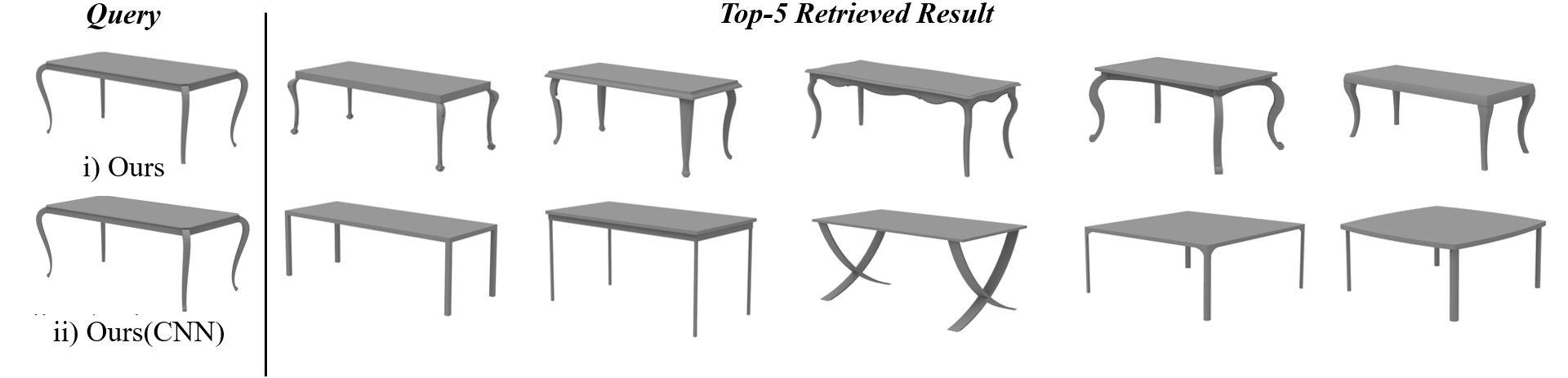}
    }
    \caption{Top-5 retrieved results among different base features and geometric feature extractors.}
    \label{fig:exp_meshcnn_vis}
\end{figure}

We also provide two examples for visual comparison. Fig.~\ref{fig:exp_meshcnn_vis}(a) is a comparison %
\YL{of different choices of base features.}
Using the %
\YL{scale-invariant}
feature as in MeshCNN~\cite{hanocka2019meshcnn}, although the semantic parts of retrieved shapes all look similar to the parts of the query shape, the size relationship \YL{is} not maintained among parts, leading to dissimilar overall shapes. In Fig.~\ref{fig:exp_meshcnn_vis}(b), a failure case of using CNN instead of our \textit{PartVAE} structure is provided. \YL{Using the classification network for feature extraction} sometimes fails to capture geometric \YL{features from thin geometry.
In contrast, with the reconstructive network of our \textit{PartVAE},}
the detailed geometric features are well-learned in the latent space of our VAE modules as shown in the experiment, showing the effectiveness of our PartVAEs.

\textbf{Variational AutoEncoder vs. AutoEncoder.}
In our observations, the part-wise differences are the key factors to distinguish between sub-classes, because the shapes of the same class share similar global structural features. That could be better modeled by VAE, because it is able to learn shape representations with such disentangled factors. We compare the performance by replacing the Variational AutoEncoder by a normal AutoEncoder in our model for validation. The performance of fine-grained retrieval of latent features learned by the VAE-based and AE-based models are given in Table~\ref{Tab:VAE_AE}. \YL{The} VAE-based approach achieves a significantly better result than AE-based approach.

\begin{table}[h]
\centering
\caption {The comparison between VAE and AE structure, $latent_{1}=64$, $latent_{2}=64$.}
\begin{tabular}{|c|c|c|c|c|c|}
\hline
            & \multicolumn{5}{c|}{micro}                                                              \\ \hline
Methods     & NN              & FT              & ST              & NDCG            & mAP             \\ \hline
AE & 0.6972          & 0.5433          & 0.6191          & 0.7457          & 0.5777          \\ \hline
VAE     & \textbf{0.7378} & \textbf{0.5670} & \textbf{0.6892} & \textbf{0.7878} & \textbf{0.6125} \\ \hline\hline
            & \multicolumn{5}{c|}{macro}                                                              \\ \hline
Methods     & NN              & FT              & ST              & NDCG            & mAP             \\ \hline
AE & 0.5248          & 0.4420          & 0.5119          & 0.6329          & 0.4801          \\ \hline
VAE     & \textbf{0.6003} & \textbf{0.4642} & \textbf{0.5631} & \textbf{0.7065} & \textbf{0.5142} \\ \hline
\end{tabular}
\label{Tab:VAE_AE}
\end{table}

\textbf{Structure information.} Finally, we compare the performance of RISA-Net with and without structural information in shape description. To remove the structural information, we just encode geometric information through the set of \textit{PartVAE}s and \textit{Part-Geo} attention mechanism, and feed their concatenation to the \textit{GlobalVAE} to form the final shape descriptor. As shown in Table~\ref{table: hierach_wo}, the shape descriptor that contains global structure information has a better performance than the shape descriptor with only geometric information.

\begin{table}[h]
\centering
\caption {The comparison between RISA-Net with and without structure information, $lr=1 \times 10^{-5}$, $latent_{1}=64$, $latent_{2}=64$.}
\label{table: hierach_wo}
\begin{tabular}{|c|c|c|c|c|c|}
\hline
            & \multicolumn{5}{c|}{micro}                                                              \\ \hline
Methods     & NN              & FT              & ST              & NDCG            & mAP             \\ \hline
PartVAE Set & 0.7038          & 0.5011          & 0.5880          & 0.7598          & 0.5754          \\ \hline
RISA-Net     & \textbf{0.7378} & \textbf{0.5670} & \textbf{0.6892} & \textbf{0.7878} & \textbf{0.6125} \\ \hline\hline
            & \multicolumn{5}{c|}{macro}                                                              \\ \hline
Methods     & NN              & FT              & ST              & NDCG            & mAP             \\ \hline
PartVAE Set & 0.5841          & 0.4092          & 0.4976          & 0.6767          & 0.4752          \\ \hline
RISA-Net     & \textbf{0.6003} & \textbf{0.4642} & \textbf{0.5631} & \textbf{0.7065} & \textbf{0.5142} \\ \hline
\end{tabular}
\end{table}

\subsection{\YL{Limitations} and Future work}
RISA-Net encodes structural information based on the spatial relationships between semantic parts, which leads to a need for correspondences between the segmented parts of the query shape and the dataset. Therefore, if the query object has totally different topology with the data in the repository, RISA-Net may fail to interpret the structural information effectively. Further research can focus on how to automatically learn the structural information and combine it with geometric information.

\section{Conclusion}
\label{sec:conclusion}
In this paper, we introduce RISA-Net, a novel framework to extract shape descriptors for fine-grained 3D object retrieval. RISA-Net is able to extract 3D shape descriptors with geometric details and global structural information, which are invariant to rigid \YL{transformation}. Trained with the attention mechanisms and the dedicatedly designed losses, RISA-Net can locate and emphasize \YL{discriminative parts}, and make a balance between structure and geometric information when representing a 3D shape. 
Thanks to the above designs, the 
Through fruitful experiments on fine-grained 3D shape retrieval, we demonstrated that RISA-Net outperforms the state of the art on fine-grained 3D object retrieval task. For future work, fine-grained sketch-based and image-based object retrieval would be a natural extension to our work. Also, with our method \YL{retrieving} fine-grained similar shapes, \YL{how to utilize the retrieved shapes for modeling of new shapes is worth also exploring}.

%% file: tex/supp.tex
\begin{center}
\vspace*{1mm}
\section*{\huge Supplementary Material}
\vspace*{5mm}
    \normalsize 
\end{center}

\section{Fine-grained 3D Object Retrieval Dataset}

Our 3D object dataset for validating fine-grained retrieval is built based on the dataset of SDM-NET\cite{gao2019sdm}. It contains 8906 3D models of six object categories, which are cars, planes, chairs, tables, guitars and knives. Each category is further manually classified into sub-classes based on the functionality and style.

\subsection{Annotation Method}
\label{sec:Annotation Method}
We adopt three steps to categorize each object category into sub-classes. Firstly, we manually divide the shapes of the same category into sub-classes according to their intra-class features, such as functionality, designation style and product model number. We leverage our classification with ShapeNet's taxonomy, and assign each sub-class with different semantic labels. A few sub-classes are assigned a label of product number by one of its representing objects. Secondly, we clean-up the dataset by discarding unrealistic models and special models that do not have similar features with any named sub-classes. Finally, we inspect the labels to make sure the preciseness of shapes under each sub-class.

\subsection{Dataset Examples}
\begin{figure*}[hb]
    \centering
    \subfigure[F1 race car]{
        \includegraphics[width=1.69in]{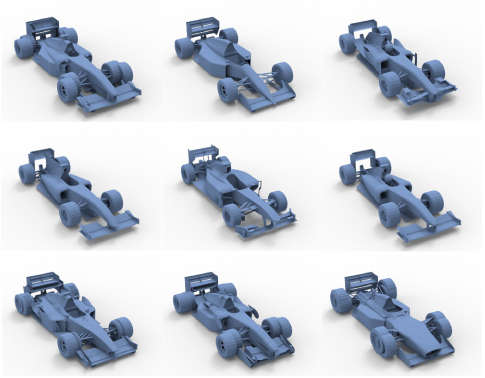}
    }
    \subfigure[crew cab pickup]{
  \includegraphics[width=1.69in]{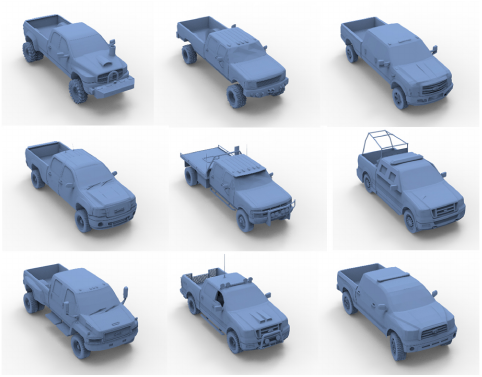}
    }
  \subfigure[convertible car]{
        \includegraphics[width=1.69in]{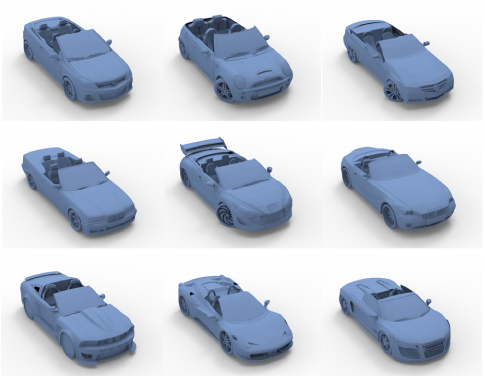}
    }
    \subfigure[2-door hatchback]{
  \includegraphics[width=1.69in]{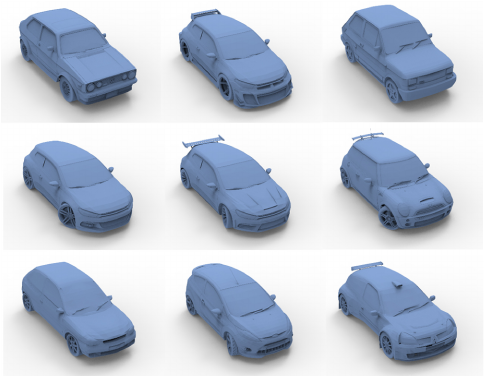}
    } 
    
    \caption{Examples of four sub-classes of cars.}
    \label{fig:example_car}
\end{figure*}

\begin{figure*}[hb]
    \centering
    \subfigure[barrel chair]{
        \includegraphics[width=2in]{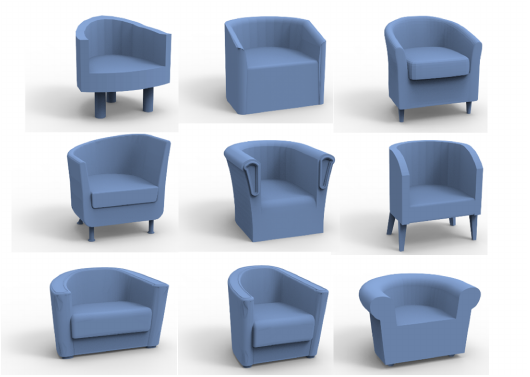}
    }
    \subfigure[X-back chair]{
  \includegraphics[width=1.59in]{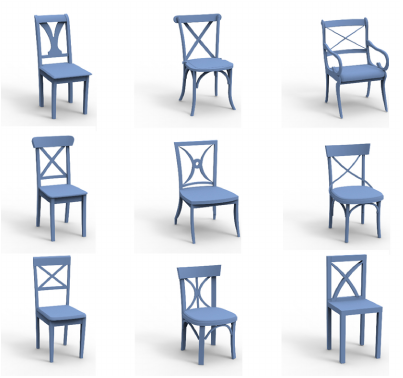}
    }
  \subfigure[splat back chair]{
        \includegraphics[width=1.59in]{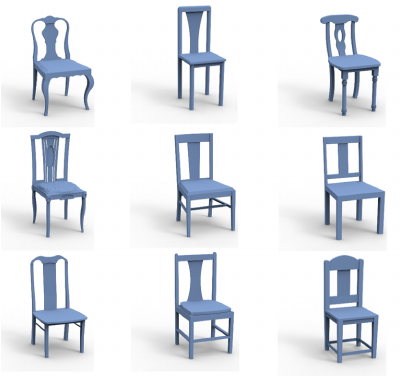}
    }
    \subfigure[banquet chair]{
  \includegraphics[width=1.59in]{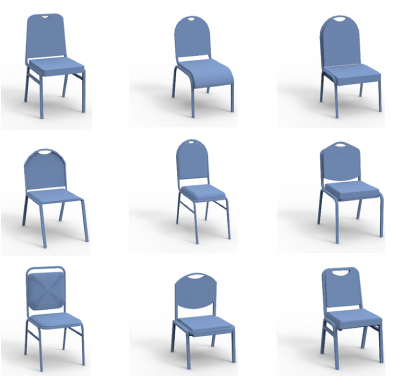}
    }  
    
    \caption{Examples of four sub-classes of chairs.}
    \label{fig:example_chair}
\end{figure*}

\begin{figure*}[hb]
    \centering
    \subfigure[combat knife]{
        \includegraphics[width=1.5in]{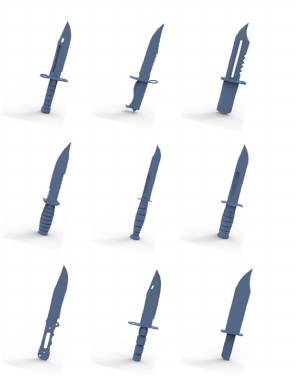}
    }
    \subfigure[chef's knife]{
  \includegraphics[width=1.5in]{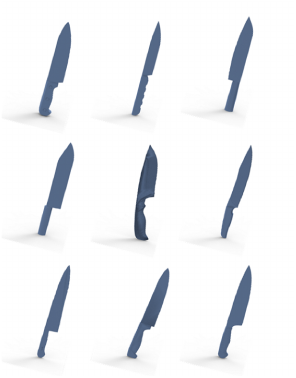}
    }
  \subfigure[machete]{
  
        \includegraphics[width=1.5in]{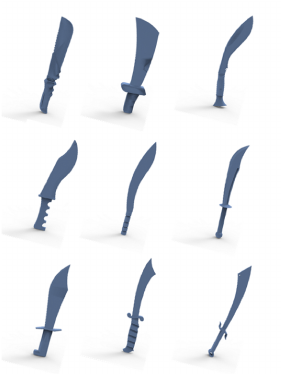}
    }
    \subfigure[foldable pocket knife]{
  \includegraphics[width=1.5in]{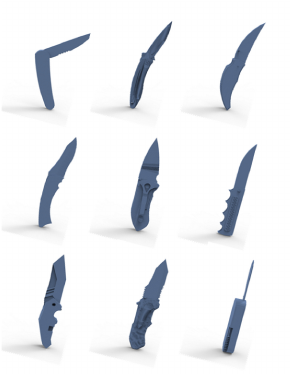}
    }  
    
    \caption{Examples of four sub-classes under the knife category.}
    \label{fig:example_knife}
\end{figure*}

\begin{figure*}[htb]
    \centering
    \subfigure[Gibson Explorer]{
        \includegraphics[width=1.5in]{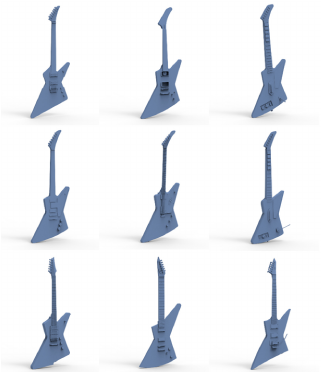}
    }
    \subfigure[Telecaster]{
  \includegraphics[width=1.5in]{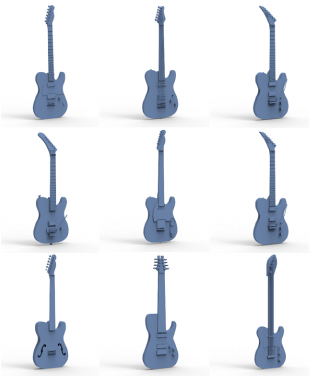}
    }
  \subfigure[Stratocaster]{
        \includegraphics[width=1.5in]{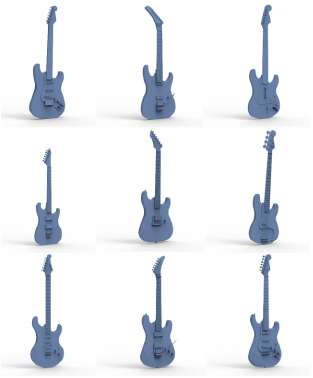}
    }
    \subfigure[cutaway guitar]{
  \includegraphics[width=1.5in]{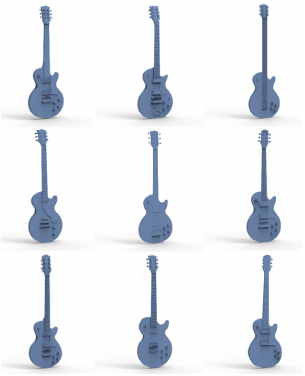}
    }  
    
    \caption{Examples of four sub-classes under guitar category}
    \label{fig:example_guitar}
\end{figure*}

\begin{figure*}[htb]
    \centering
    \subfigure[Car]{
        \includegraphics[width=1.5in]{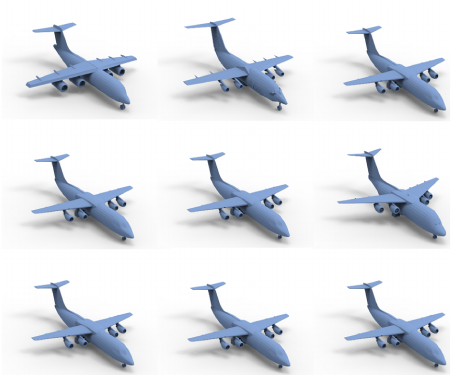}
    }
    \subfigure[Plane]{
	\includegraphics[width=1.5in]{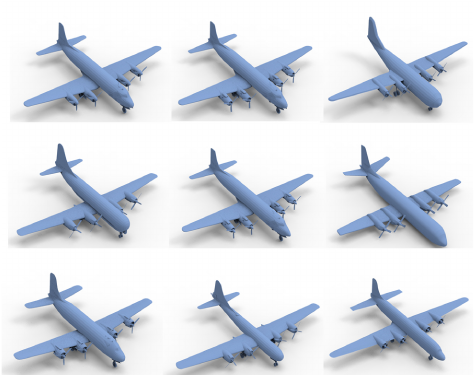}
    }
  \subfigure[MIG15]{
        \includegraphics[width=1.5in]{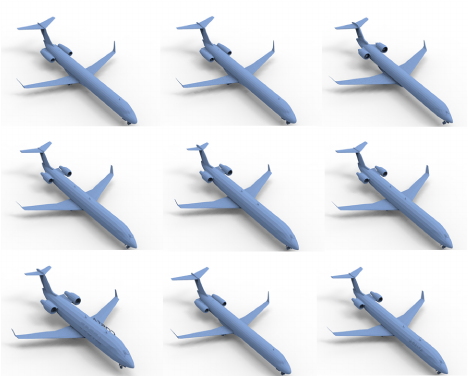}
    }
    \subfigure[DC10]{
	\includegraphics[width=1.5in]{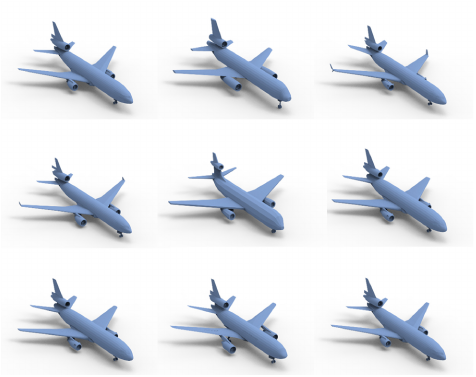}
    }  
    
    \caption{Examples of four sub-classes under the plane category.}
    \label{fig:example_plane}
\end{figure*}

\begin{figure*}[htb]
    \centering
    \subfigure[simple round coffee table]{
        \includegraphics[width=1.55in]{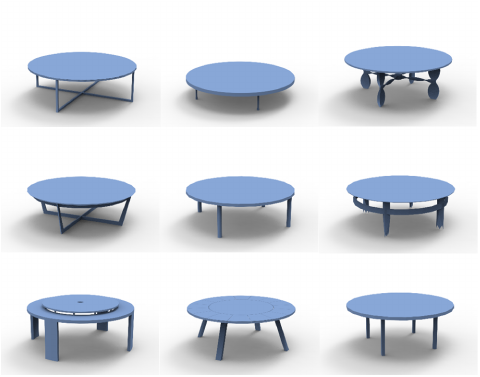}
    }
    \subfigure[L-shape desk]{
	\includegraphics[width=1.55in]{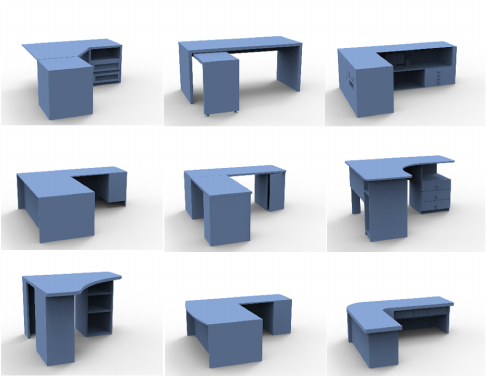}
    }
  \subfigure[classic rectangular end table]{
        \includegraphics[width=1.4in]{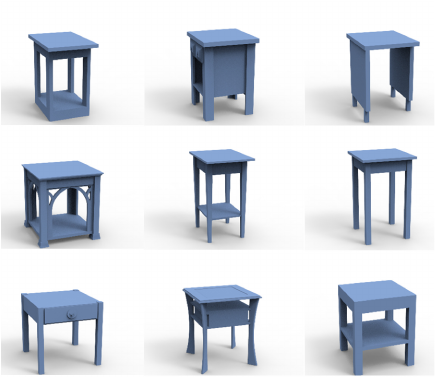}
    }
    \subfigure[office desk]{
	\includegraphics[width=1.55in]{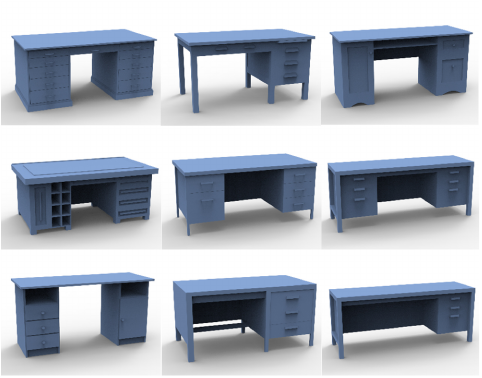}
    }  
    
    \caption{Examples of four sub-classes under the table category.}
    \label{fig:example_table}
\end{figure*}

Figure~\ref{fig:example_car} - Figure~\ref{fig:example_table} illustrate several examples of some sub-classes of each category. Note that objects of each sub-class share similar high-level structural features, but have distinguishable part-wise features. For example, in Figure~\ref{fig:example_car}, the four sub-classes of cars have similar overall structures, but cars of the sub-class ``F1 race car'' are single-seated, open cockpit racing cars, and cars that are labeled by ``crew cab pickup'' are pickups with a single row of seats. %

\begin{table*}[ht]
  \centering
  \caption{Statistics of sub-classes of 3 categories.}
  \setlength{\tabcolsep}{0.8mm}{
    \begin{tabular}{lr|rrrr|rrrr}
    \toprule
    \multicolumn{2}{c|}{\textbf{Car}} & \multicolumn{4}{c|}{\textbf{Plane}} & \multicolumn{4}{c}{\textbf{Chair}} \\
    \midrule
    \midrule
    Sub-Class & \multicolumn{1}{l|}{Num} & \multicolumn{1}{l}{Sub-Class} & \multicolumn{1}{l|}{Num} & \multicolumn{1}{l}{Sub-Class} & \multicolumn{1}{l|}{Num} & \multicolumn{1}{l}{Sub-Class} & \multicolumn{1}{l|}{Num} & \multicolumn{1}{l}{Sub-Class} & \multicolumn{1}{l}{Num} \\
    \midrule
    sedan & 219   & \multicolumn{1}{l}{A320} & \multicolumn{1}{r|}{181} & \multicolumn{1}{l}{BAe 146} & 16    & \multicolumn{1}{l}{tucker} & \multicolumn{1}{r|}{314} & \multicolumn{1}{l}{elbow chair} & 17 \\
    sports car & 140   & \multicolumn{1}{l}{B747} & \multicolumn{1}{r|}{158} & \multicolumn{1}{l}{B737-200} & 15    & \multicolumn{1}{l}{straight chair} & \multicolumn{1}{r|}{251} & \multicolumn{1}{l}{x-back } & 16 \\
    convertible cars & 106   & \multicolumn{1}{l}{CRJ} & \multicolumn{1}{r|}{153} & \multicolumn{1}{l}{A300-600} & 15    & \multicolumn{1}{l}{conference chair} & \multicolumn{1}{r|}{201} & \multicolumn{1}{l}{outdoor lounge} & 14 \\
    extended cab pickup & 100   & \multicolumn{1}{l}{B737-800} & \multicolumn{1}{r|}{142} & \multicolumn{1}{l}{B767-300w} & 13    & \multicolumn{1}{l}{slat chair} & \multicolumn{1}{r|}{146} & \multicolumn{1}{l}{sofa} & 13 \\
    crew cab pickup & 93    & \multicolumn{1}{l}{B777-300} & \multicolumn{1}{r|}{89} & \multicolumn{1}{l}{E170} & 12    & \multicolumn{1}{l}{lounge chair} & \multicolumn{1}{r|}{116} & \multicolumn{1}{l}{bench} & 13 \\
    supercar & 85    & \multicolumn{1}{l}{DC10} & \multicolumn{1}{r|}{84} & \multicolumn{1}{l}{Lockheed F117} & 11    & \multicolumn{1}{l}{prouve chair} & \multicolumn{1}{r|}{113} & \multicolumn{1}{l}{norman chair} & 13 \\
    2-door hatchback & 83    & \multicolumn{1}{l}{A340-600} & \multicolumn{1}{r|}{79} & \multicolumn{1}{l}{E195} & 10    & \multicolumn{1}{l}{sqaure back chair} & \multicolumn{1}{r|}{102} & \multicolumn{1}{l}{barrel chair} & 11 \\
    coupe & 80    & \multicolumn{1}{l}{A330-300} & \multicolumn{1}{r|}{78} & \multicolumn{1}{l}{Concorde} & 10    & \multicolumn{1}{l}{parson chair} & \multicolumn{1}{r|}{98} & \multicolumn{1}{l}{arne jacobsen egg chair} & 11 \\
    vintage muscle car & 72    & \multicolumn{1}{l}{B787-800} & \multicolumn{1}{r|}{67} & \multicolumn{1}{l}{MIG15} & 10    & \multicolumn{1}{l}{cantilever side chair} & \multicolumn{1}{r|}{90} & \multicolumn{1}{l}{cantilever lounge chair} & 11 \\
    minivan & 55    & \multicolumn{1}{l}{B727} & \multicolumn{1}{r|}{67} & \multicolumn{1}{l}{B737-600} & 9     & \multicolumn{1}{l}{sled base chair} & \multicolumn{1}{r|}{68} & \multicolumn{1}{l}{bentwood chairs} & 11 \\
    regular cab pickup & 43    & \multicolumn{1}{l}{Lockheed Martin X-35} & \multicolumn{1}{r|}{60} & \multicolumn{1}{l}{Cessna C208} & 9     & \multicolumn{1}{l}{ladder back} & \multicolumn{1}{r|}{65} & \multicolumn{1}{l}{banquet chair} & 10 \\
    4-door hatchback & 35    & \multicolumn{1}{l}{middle wing propeller} & \multicolumn{1}{r|}{54} & \multicolumn{1}{l}{Super Transport} & 4     & \multicolumn{1}{l}{high stool chair} & \multicolumn{1}{r|}{62} & \multicolumn{1}{l}{park bench} & 10 \\
    coach bus & 29    & \multicolumn{1}{l}{A380} & \multicolumn{1}{r|}{54} &       &       & \multicolumn{1}{l}{club chair} & \multicolumn{1}{r|}{56} & \multicolumn{1}{l}{windsor chair} & 9 \\
    f1 race car & 29    & \multicolumn{1}{l}{B767-300} & \multicolumn{1}{r|}{48} &       &       & \multicolumn{1}{l}{vintage french} & \multicolumn{1}{r|}{55} & \multicolumn{1}{l}{eiffel aimchair} & 8 \\
    convertible vintage & 27    & \multicolumn{1}{l}{B777-200} & \multicolumn{1}{r|}{47} &       &       & \multicolumn{1}{l}{revolving chair} & \multicolumn{1}{r|}{52} & \multicolumn{1}{l}{chiavari chairs} & 7 \\
    luxury car & 25    & \multicolumn{1}{l}{high wing propeller} & \multicolumn{1}{r|}{44} &       &       & \multicolumn{1}{l}{splat back chair} & \multicolumn{1}{r|}{48} & \multicolumn{1}{l}{industrial café} & 6 \\
    cargo van & 19    & \multicolumn{1}{l}{B757-200} & \multicolumn{1}{r|}{42} &       &       & \multicolumn{1}{l}{adirondack} & \multicolumn{1}{r|}{35} & \multicolumn{1}{l}{easter egg chair} & 5 \\
    off-roader & 19    & \multicolumn{1}{l}{P51 Musang} & \multicolumn{1}{r|}{36} &       &       & \multicolumn{1}{l}{navy chair} & \multicolumn{1}{r|}{35} & \multicolumn{1}{l}{outdoor iron chair} & 5 \\
    muv   & 17    & \multicolumn{1}{l}{B737-832} & \multicolumn{1}{r|}{34} &       &       & \multicolumn{1}{l}{folding chair} & \multicolumn{1}{r|}{32} &       &  \\
    cargo truck & 16    & \multicolumn{1}{l}{B707} & \multicolumn{1}{r|}{28} &       &       & \multicolumn{1}{l}{task chair} & \multicolumn{1}{r|}{31} &       &  \\
    suv   & 16    & \multicolumn{1}{l}{L-1011 Tristar} & \multicolumn{1}{r|}{21} &       &       & \multicolumn{1}{l}{qing chair} & \multicolumn{1}{r|}{29} &       &  \\
    truck tractor & 15    & \multicolumn{1}{l}{B737-2VG} & \multicolumn{1}{r|}{19} &       &       & \multicolumn{1}{l}{panton chair} & \multicolumn{1}{r|}{27} &       &  \\
    all-terrain vehicle & 13    & \multicolumn{1}{l}{B757-200w} & \multicolumn{1}{r|}{19} &       &       & \multicolumn{1}{l}{eiffel chair} & \multicolumn{1}{r|}{22} &       &  \\
    streched limo & 12    & \multicolumn{1}{l}{B737-300} & \multicolumn{1}{r|}{18} &       &       & \multicolumn{1}{l}{drafting chair} & \multicolumn{1}{r|}{20} &       &  \\
    streched suv & 7     & \multicolumn{1}{l}{E190} & \multicolumn{1}{r|}{17} &       &       & \multicolumn{1}{l}{slipper chair} & \multicolumn{1}{r|}{19} &       &  \\
    semi-trailer truck & 6     & \multicolumn{1}{l}{Rafale} & \multicolumn{1}{r|}{17} &       &       & \multicolumn{1}{l}{barcelona chair} & \multicolumn{1}{r|}{18} &       &  \\
    school bus & 4     & \multicolumn{1}{l}{MIRAGE 2000} & \multicolumn{1}{r|}{17} &       &       & \multicolumn{1}{l}{rocking chair} & \multicolumn{1}{r|}{17} &       &  \\
    \midrule
    Total & 1365  & \multicolumn{4}{r|}{1807}     & \multicolumn{4}{r}{2312} \\
    \bottomrule
    \end{tabular}}%
  \label{tab:dist1}%
\end{table*}%

\begin{table*}[hb]
  \centering
  \caption{Statistics of sub-classes of 3 categories.}
  \setlength{\tabcolsep}{0.5mm}{
\begin{tabular}{lr|rr|rrrr}
    \toprule
    \multicolumn{2}{c|}{\textbf{Knife}} & \multicolumn{2}{c|}{\textbf{Guitar}} & \multicolumn{4}{c}{\textbf{Table}} \\
    \midrule
    \midrule
    Sub-Class & \multicolumn{1}{l|}{Num} & \multicolumn{1}{l}{Sub-Class} & \multicolumn{1}{l|}{Num} & \multicolumn{1}{l}{Sub-Class} & \multicolumn{1}{l|}{Num} & \multicolumn{1}{l}{Sub-Class} & \multicolumn{1}{l}{Num} \\
    sword & 54    & \multicolumn{1}{l}{Stratocaster} & 153   & \multicolumn{1}{l}{modern rectangualr dining table} & \multicolumn{1}{r|}{259} & \multicolumn{1}{l}{frame rectanglular dinner table} & 49 \\
    combat knife & 37    & \multicolumn{1}{l}{Gibson Flying V} & 89    & \multicolumn{1}{l}{classic rectangualr end table } & \multicolumn{1}{r|}{192} & \multicolumn{1}{l}{pool table} & 40 \\
    machete & 21    & \multicolumn{1}{l}{acoustic guitar} & 78    & \multicolumn{1}{l}{pedestal round end table} & \multicolumn{1}{r|}{173} & \multicolumn{1}{l}{pedestal rectangualr end table} & 40 \\
    foldable pocket knife & 14    & \multicolumn{1}{l}{Les Paul} & 67    & \multicolumn{1}{l}{pedestal round dinning table } & \multicolumn{1}{r|}{130} & \multicolumn{1}{l}{pedestal rectanglular dinner table} & 38 \\
    dagger & 12    & \multicolumn{1}{l}{Widow} & 55    & \multicolumn{1}{l}{simple rectangualr coffee table} & \multicolumn{1}{r|}{128} & \multicolumn{1}{l}{pedestal round coffee table } & 35 \\
    chef's knife & 10    & \multicolumn{1}{l}{SG guitar} & 50    & \multicolumn{1}{l}{simple rectangualr dining table} & \multicolumn{1}{r|}{110} & \multicolumn{1}{l}{writing desk} & 32 \\
    throwing knife & 9     & \multicolumn{1}{l}{Star} & 32    & \multicolumn{1}{l}{console table} & \multicolumn{1}{r|}{100} & \multicolumn{1}{l}{office table} & 30 \\
    cleaver & 8     & \multicolumn{1}{l}{Gibson Explorer} & 31    & \multicolumn{1}{l}{traditional rectanglular dinner table} & \multicolumn{1}{r|}{100} & \multicolumn{1}{l}{ping-pong table} & 29 \\
    dinner knife & 8     & \multicolumn{1}{l}{cutaway} & 27    & \multicolumn{1}{l}{simple round end table} & \multicolumn{1}{r|}{100} & \multicolumn{1}{l}{L-shape desk} & 28 \\
    karambit & 7     & \multicolumn{1}{l}{Telecaster} & 21    & \multicolumn{1}{l}{simple round dinning table } & \multicolumn{1}{r|}{90} & \multicolumn{1}{l}{cross style round end table } & 24 \\
    dragon knife & 6     & \multicolumn{1}{l}{Beast} & 16    & \multicolumn{1}{l}{modern rectangualr coffee table} & \multicolumn{1}{r|}{85} & \multicolumn{1}{l}{cabinet} & 20 \\
          &       & \multicolumn{1}{l}{double-neck guitar} & 10    & \multicolumn{1}{l}{traditional rectangle coffee table} & \multicolumn{1}{r|}{85} & \multicolumn{1}{l}{cross style round dinning table } & 20 \\
          &       &       &       & \multicolumn{1}{l}{office desk} & \multicolumn{1}{r|}{80} & \multicolumn{1}{l}{cart} & 17 \\
          &       &       &       & \multicolumn{1}{l}{simple round coffee table } & \multicolumn{1}{r|}{77} & \multicolumn{1}{l}{hourglass round dinning table} & 14 \\
          &       &       &       & \multicolumn{1}{l}{training table} & \multicolumn{1}{r|}{74} & \multicolumn{1}{l}{desk shell} & 12 \\
          &       &       &       & \multicolumn{1}{l}{TV stand} & \multicolumn{1}{r|}{67} & \multicolumn{1}{l}{secretaire} & 12 \\
          &       &       &       & \multicolumn{1}{l}{multi-media table} & \multicolumn{1}{r|}{64} & \multicolumn{1}{l}{hourglass round dinner table} & 12 \\
          &       &       &       & \multicolumn{1}{l}{frame rectanglular coffee table} & \multicolumn{1}{r|}{57} & \multicolumn{1}{l}{lectern} & 9 \\
          &       &       &       & \multicolumn{1}{l}{conference table} & \multicolumn{1}{r|}{56} & \multicolumn{1}{l}{picnic table} & 8 \\
          &       &       &       & \multicolumn{1}{l}{work bench} & \multicolumn{1}{r|}{52} & \multicolumn{1}{l}{cross style round coffee table } & 8 \\
          &       &       &       & \multicolumn{1}{l}{X-base rectanglular dinner table} & \multicolumn{1}{r|}{51} &       &  \\
    \midrule
    Total & 186   & \multicolumn{2}{r|}{629} &       & \multicolumn{3}{r}{2607} \\
    \bottomrule
    \end{tabular}}%
  \label{tab:dist2}%
\end{table*}%

\subsection{Sub-class Distribution}
\begin{figure*}[htbp]
    \centering
    \subfigure[car]{
        \includegraphics[width=2.75in]{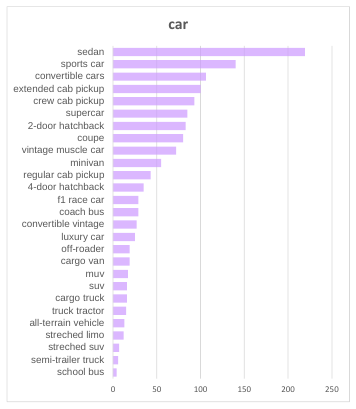}
    }
    \subfigure[plane]{
	\includegraphics[width=2.5in]{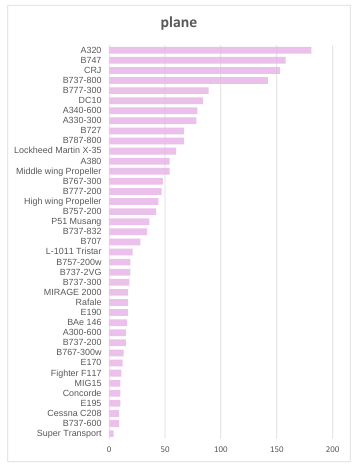}
    }
    \subfigure[chair]{
    \includegraphics[width=2.5in]{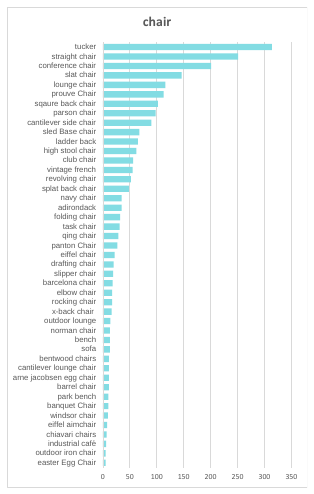}
    }
    \subfigure[table]{
	\includegraphics[width=2.75in]{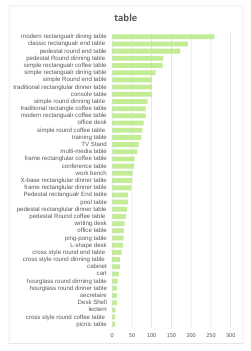}
    }
    \subfigure[guitar]{
        \includegraphics[width=2.7in]{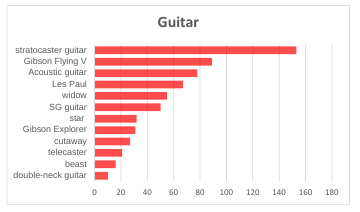}
    }
    \subfigure[knife]{
	\includegraphics[width=2.5in]{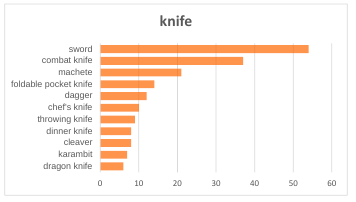}
    }
    \caption{Data Distribution}
    \label{fig:dist3}
\end{figure*}

Table~\ref{tab:dist1}-Table~\ref{tab:dist2} provide the number of shapes under each sub-class on six object categories. Fig.~\ref{fig:dist3} illustrate the sub-class distribution, ranked by the number of models from high to low.

\newpage
\clearpage

\section{Retrieval Results}

We use a shape repository containing both train and test set to evaluate its performance on large-scale dataset retrieval. In Table~\ref{Tab:all_table}, we compare RISA-Net with other methods using five metrics on six object categories. RISA-net outperforms other methods on large-scale fin-grained shape retrieval. 

We provide top-10 retrieval results when retrieving shapes from 6 object categories in our fine-grained 3D shape retrieval dataset. The query shape is from test set, and shape repository contains all shape of the same sub-class.
The result shows that RISA-Net not only retrieves models with similar global structure, but also retrieves model with similar geometric features. 

\begin{table*}[h]
\caption{Comparison between RISA-Net and other methods using five metrics on six object categories. The query shape is from test set, and shape repository contains all shape of the same sub-class.}
\vspace{.1cm}
\centering
\begin{tabular}{|c|c|c|c|c|c|c|c|c|c|c|}
\hline
\multicolumn{1}{|l|}{} & \multicolumn{5}{c|}{micro} & \multicolumn{5}{c|}{macro} \\ \hline
 Methods               & NN              & FT              & ST              & NDCG            & mAP             & NN              & FT              & ST              & NDCG            & mAP             \\ \hline
SHD~\cite{kazhdan2003rotation}               & 0.1983          & 0.1532          & 0.2749          & 0.5514          & 0.1529          & 0.1316          & 0.1013          & 0.1781          & 0.4604          & 0.1026          \\
                          LFD~\cite{chen2003visual}                  & 0.2662          & 0.1947          & 0.3395          & 0.5835          & 0.1949          & 0.1717          & 0.1225          & 0.2147          & 0.4870          & 0.1263          \\
                          MVCNN~\cite{su2015multi}                    & 0.4346          & 0.1718          & 0.2754          & 0.5878          & 0.1635          & 0.3376          & 0.1246          & 0.1914          & 0.4995          & 0.1150          \\
                          RotationNet~\cite{thomas2018tensor}                & 0.1569          & 0.1371          & 0.2378          & 0.5304          & 0.1501          & 0.1274          & 0.1033          & 0.1687          & 0.4501          & 0.1146          \\
                          Spherical~\cite{esteves2018learning}               & 0.2399          & 0.1301          & 0.2231          & 0.5353          & 0.1352          & 0.2233          & 0.1085          & 0.1807          & 0.4833          & 0.1044          \\
                          SVSL~\cite{han2018seqviews2seqlabels}                   & 0.3076          & 0.2309          & 0.3704          & 0.6134          & 0.2370          & 0.1955          & 0.1437          & 0.2372          & 0.5102          & 0.1508          \\
                          MeshCNN~\cite{hanocka2019meshcnn}               & 0.4656          & 0.3070          & 0.4667          & 0.6681          & 0.3061          & 0.3453          & 0.2451          & 0.3824          & 0.5635          & 0.2519          \\                          
                          Ours                  & \textbf{0.7831} & \textbf{0.5651} & \textbf{0.6544} & \textbf{0.8101} & \textbf{0.5817} & \textbf{0.6908} & \textbf{0.4672} & \textbf{0.5401} & \textbf{0.7385} & \textbf{0.4817} \\\hline
\end{tabular}
\label{Tab:all_table}
\end{table*}

\begin{figure*}[h]
\centering
\includegraphics[scale = 0.7]{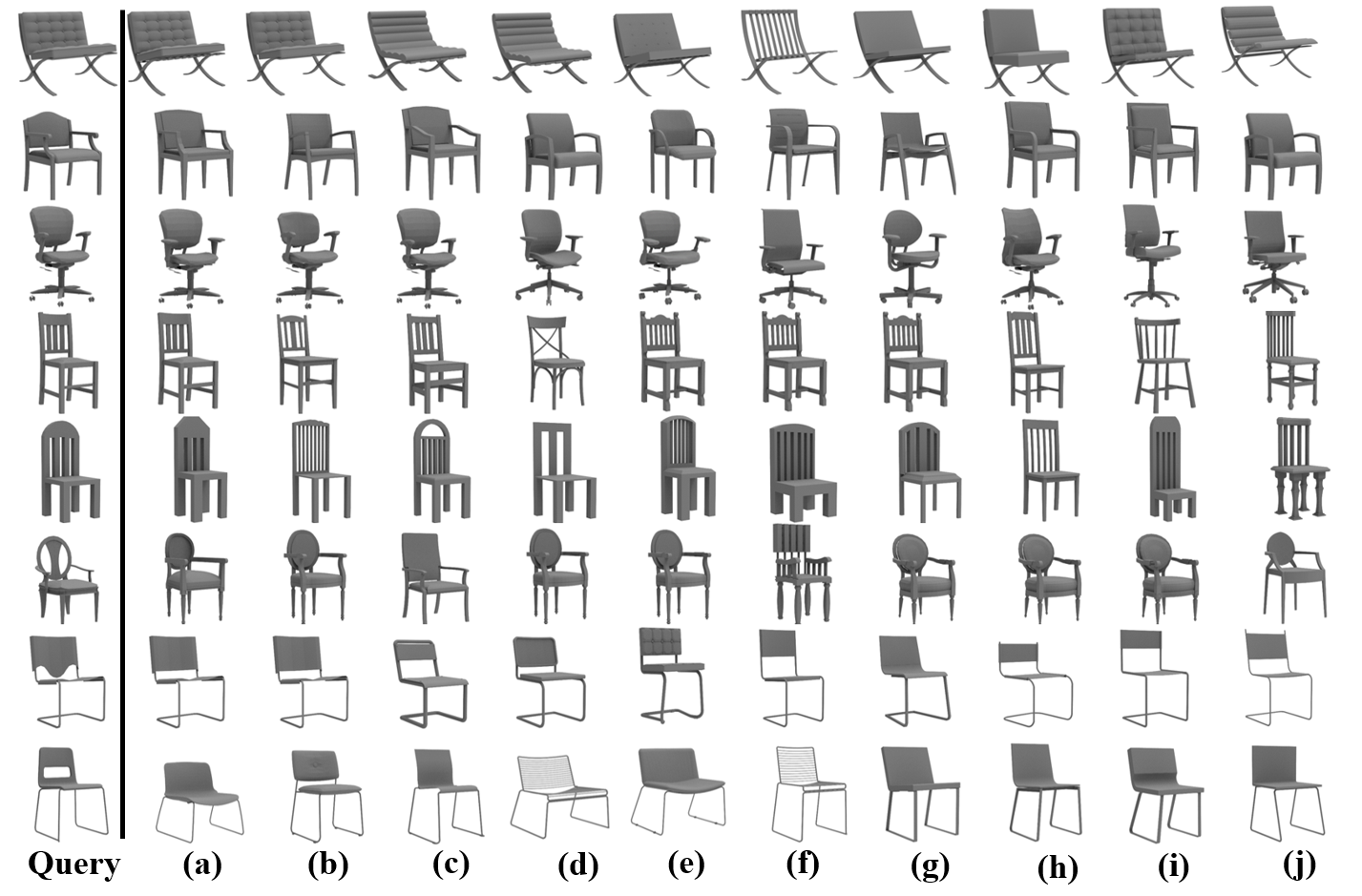}
   \caption{Retrieval Results of Chairs}
   \label{fig:sup_result_chair}
\end{figure*}

\begin{figure*}[h]
\centering
\includegraphics[scale = 0.6]{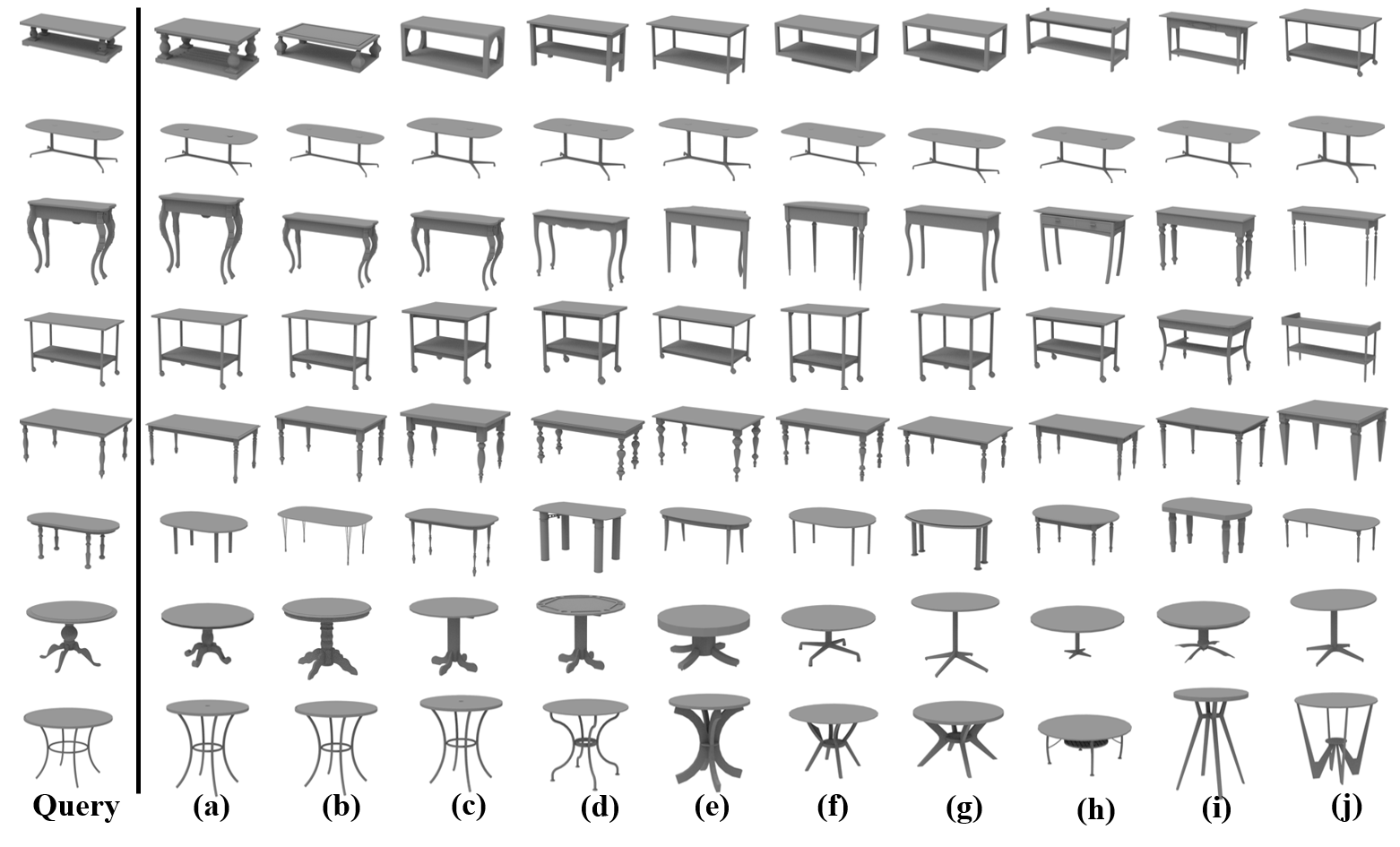}
  \caption{Retrieval Results of Tables}
  \label{fig:sup_result_table}
\end{figure*}

\begin{figure*}[h]
\centering
\includegraphics[scale = 0.6]{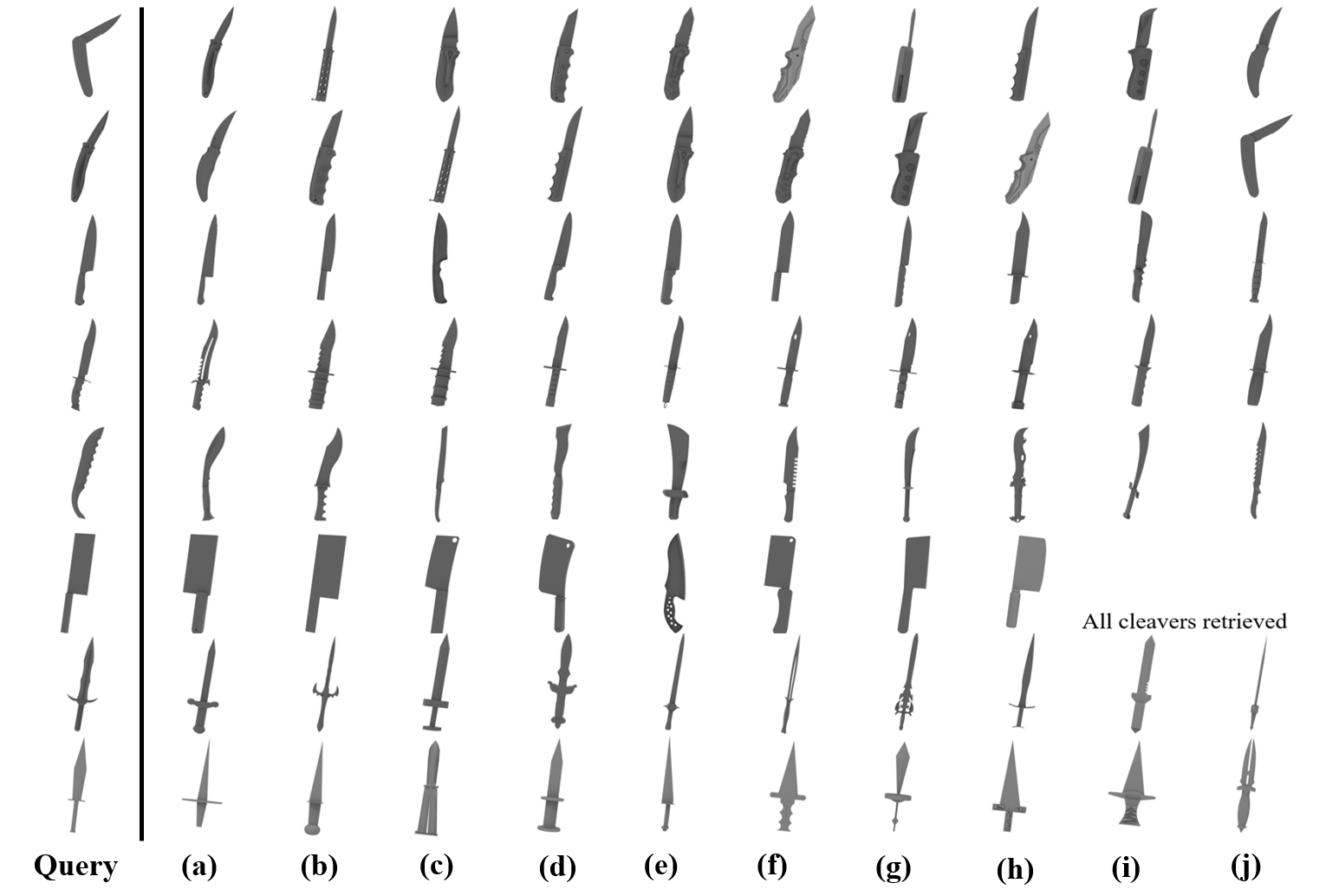}
  \caption{Retrieval Results of Knives}
  \label{fig:sup_result_knife}
\end{figure*}

\begin{figure*}[h]
\centering
\includegraphics[scale = 0.6]{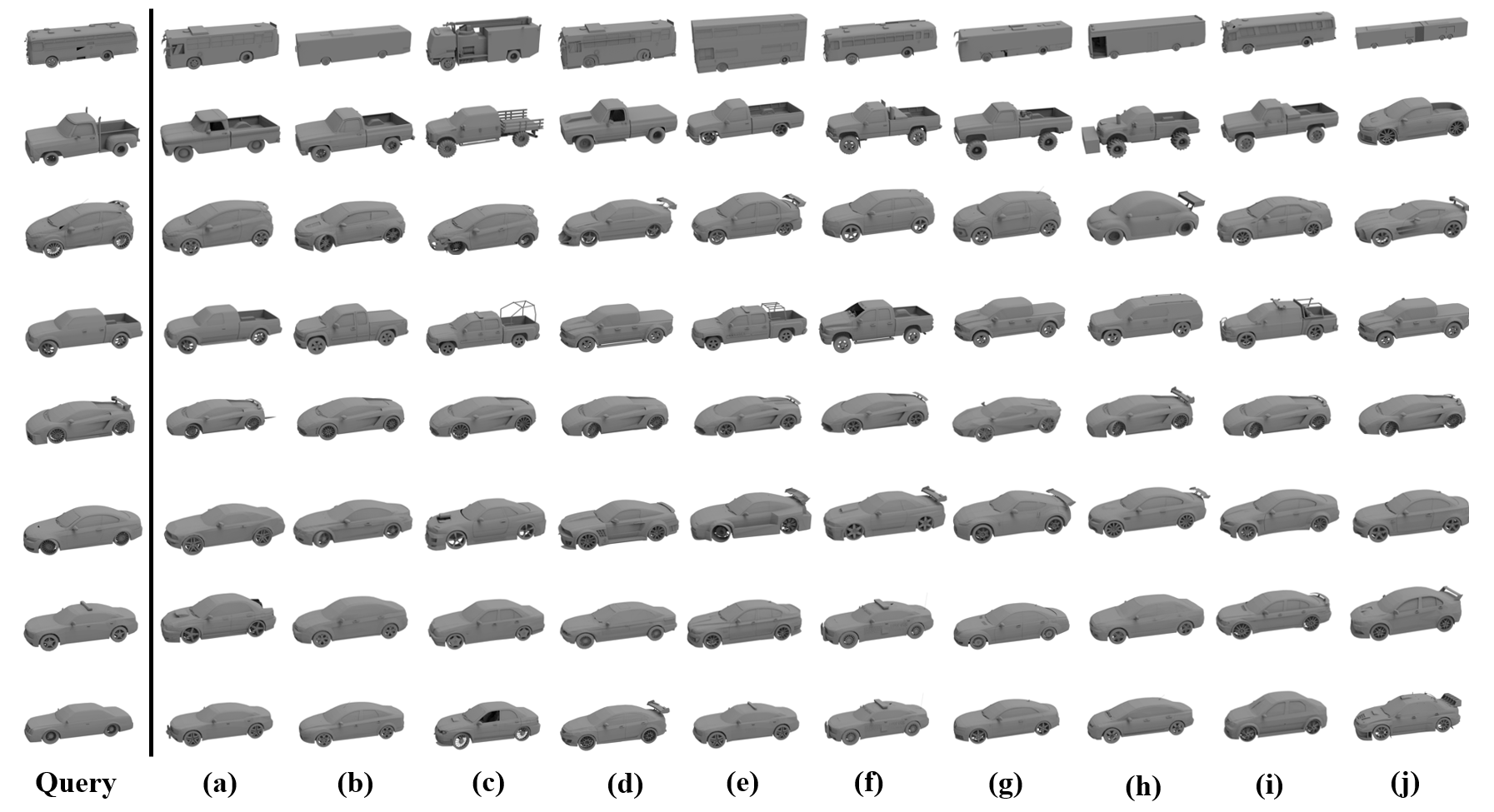}
  \caption{Retrieval Results of Cars}
  \label{fig:sup_result_car}
\end{figure*}

\begin{figure*}[h]
\centering
\includegraphics[scale = 0.6]{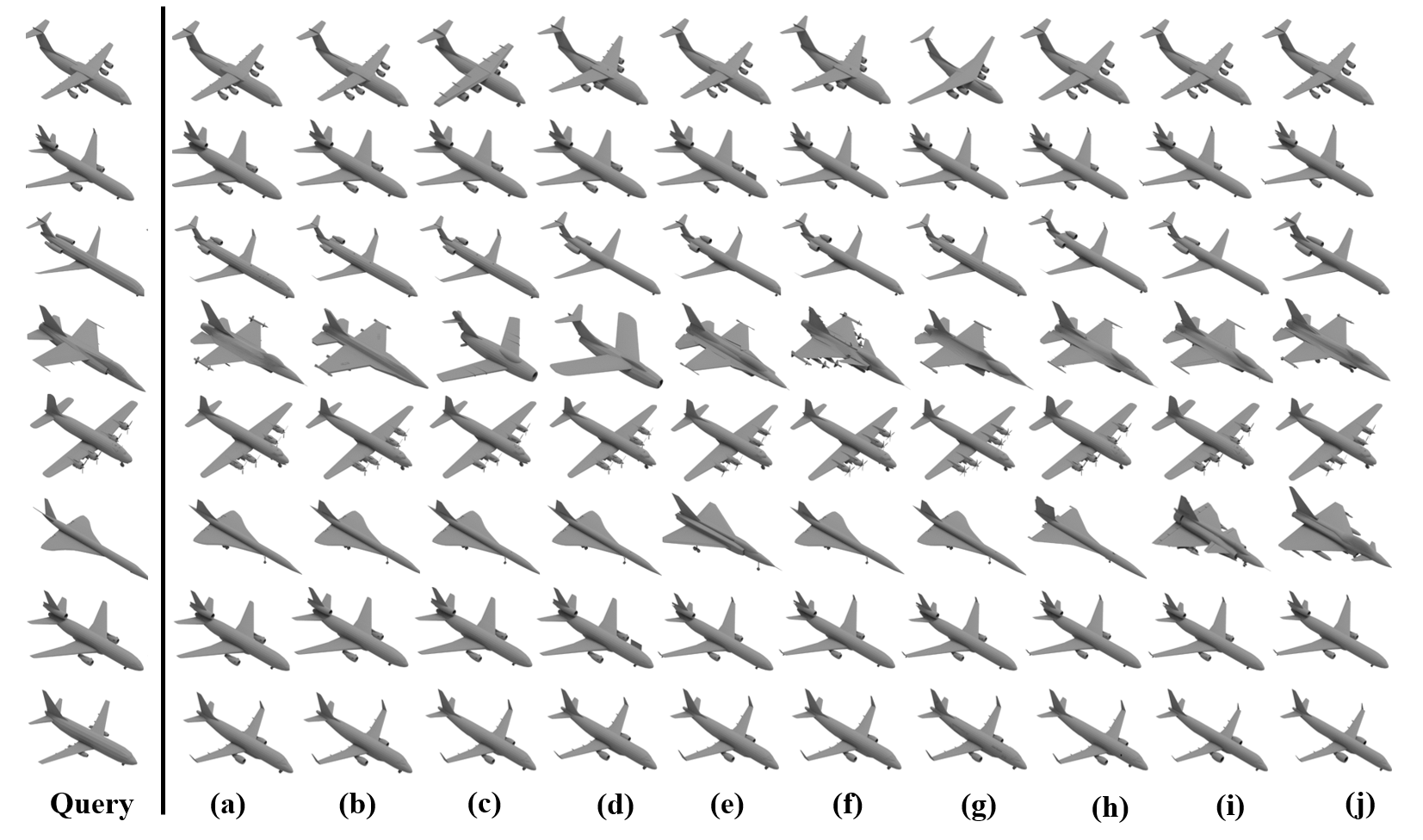}
  \caption{Retrieval Results of Planes}
  \label{fig:sup_result_plane}
\end{figure*}

\begin{figure*}[t]
\centering
\includegraphics[scale = 0.75]{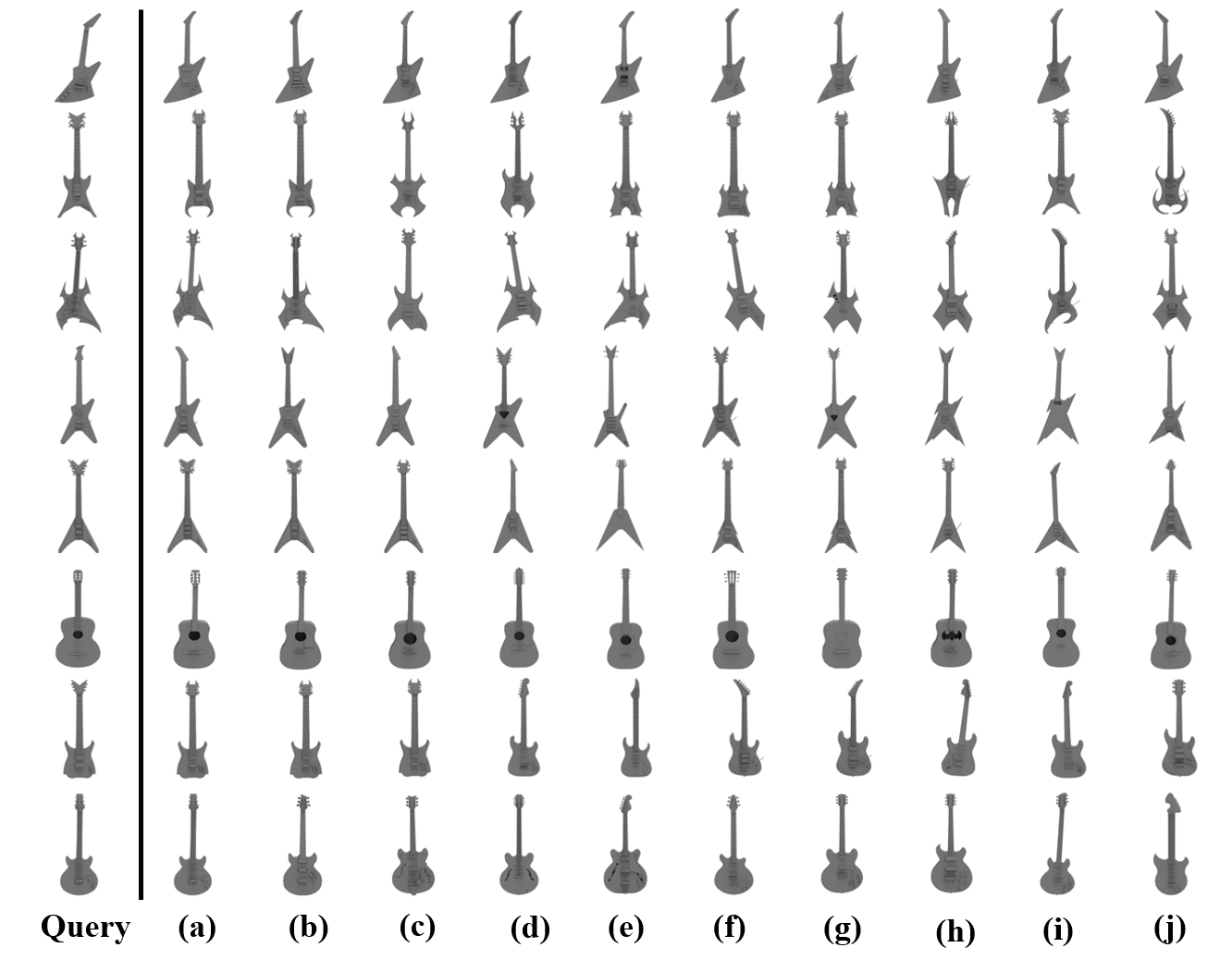}
  \caption{Retrieval Results of Guitars}
  \label{fig:sup_result_guitar}
\end{figure*}